\pdfoutput=1

\documentclass[11pt]{article}
\usepackage{acl}

\usepackage{times}
\usepackage{latexsym}
\usepackage[T1]{fontenc}
\usepackage[utf8]{inputenc}
\usepackage{microtype}
\usepackage{inconsolata}
\usepackage{graphicx}
\usepackage{amsmath}
\usepackage{booktabs}
\usepackage{lipsum}
\usepackage[caption=false]{subfig}
\usepackage{float}
\usepackage{multirow}
\usepackage{amsmath}
\usepackage{changepage}
\usepackage{natbib}
\usepackage{amsmath}
\usepackage{stmaryrd}
\usepackage{dsfont}
\usepackage{geometry} 
\usepackage{changepage}
\usepackage{makecell}
\usepackage{listings} 
\usepackage{framed}
\usepackage{amsfonts}
\usepackage{algorithm}
\usepackage{algorithmicx}
\usepackage{algpseudocode}
\usepackage{relsize}
\usepackage{tcolorbox}

\usepackage[]{todonotes}                                 
\makeatletter
\newcommand*\iftodonotes{\if@todonotes@disabled\expandafter\@secondoftwo\else\expandafter\@firstoftwo\fi}  
\makeatother

\newcommand{\rsatwo}{\textbf{(RSA)}\textsuperscript{2}}
\newcommand{\rsatwospace}{\rsatwo\,\,}
\newcommand{\rsatwolearned}{\textbf{RSC--RSA}}
\newcommand{\rsatwolearnedspace}{\rsatwolearned\,\,}

\algnewcommand{\Hypothesis}[1]{\textbf{Hypothesis:} #1}
\title{\rsatwo: A Rhetorical-Strategy-Aware Rational Speech Act Framework for Figurative Language Understanding} 

\author{%
    \bf \textsuperscript{*}Cesare Spinoso-Di Piano\textsuperscript{1,2} \quad\textsuperscript{*}David Austin\textsuperscript{1,2} \\
    \bf \textbf{Pablo Piantanida}\textsuperscript{2,3,4} \quad \textbf{Jackie Chi Kit Cheung}\textsuperscript{1,2,5} \vspace{1mm}\\
    \textsuperscript{1}McGill University, \quad \textsuperscript{2}Mila - Quebec AI Institute \\
    \textsuperscript{3}International Laboratory on Learning Systems (ILLS)\\ 
    \textsuperscript{4}CNRS, CentraleSup\'elec -  Université Paris-Saclay, \quad \textsuperscript{5}Canada CIFAR AI Chair, Mila \vspace{1mm}\\
    \texttt{\{cesare.spinoso,david.austin,pablo.piantanida,cheungja\}@mila.quebec}\\
}

\begin{document}

\maketitle

\begin{abstract}
        Figurative language (e.g., irony, hyperbole, understatement) is ubiquitous in human communication, resulting in utterances where the literal and the intended meanings do not match. The Rational Speech Act (RSA) framework, which explicitly models speaker intentions, is the most widespread theory of probabilistic pragmatics, but existing implementations are either unable to account for figurative expressions or require modeling the implicit motivations for using figurative language (e.g., to express joy or annoyance) in a setting-specific way. In this paper, we introduce the \textbf{R}hetorical-\textbf{S}trategy-\textbf{A}ware \textbf{RSA} \rsatwospace framework which models figurative language use by considering a speaker's employed rhetorical strategy. We show that \rsatwospace enables human-compatible interpretations of non-literal utterances without modeling a speaker's motivations for being non-literal. Combined with LLMs, it achieves state-of-the-art performance on the ironic split of PragMega+, a new irony interpretation dataset introduced in this study.\footnote{Code and data available at \url{https://github.com/cesare-spinoso/rsa2}.} 
\end{abstract}

\def\thefootnote{*}\footnotetext{Equal contribution.}\def\thefootnote{\arabic{footnote}}

\section{Introduction}

\label{sec:intro}

\begin{figure}[t]
    \centering
    \includegraphics[width=\linewidth]{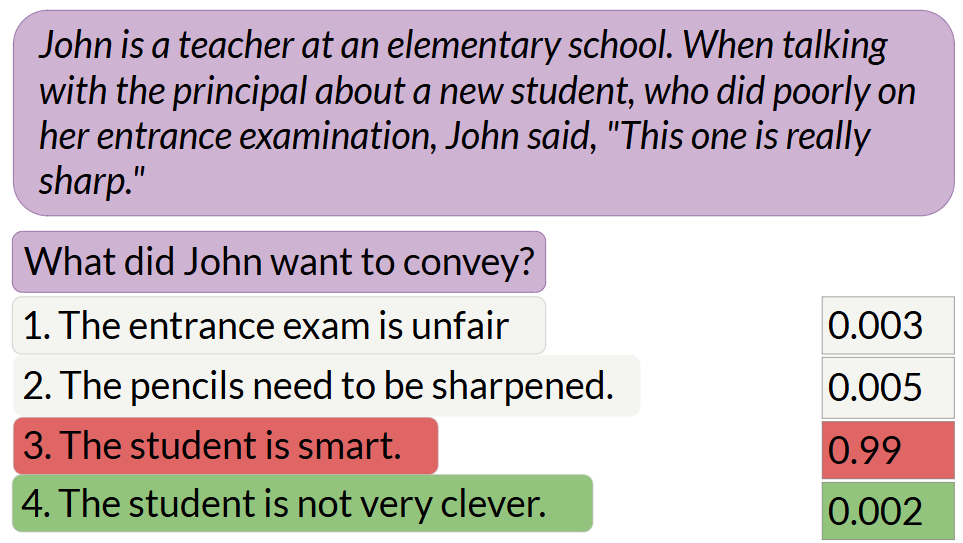}
    \caption{A sample from the PragMega dataset \citep{hu-etal-2023-fine} showing probabilities of intended meanings for a given scenario. The probabilities are computed using the Mistral-7B-Instruct model (averaged across 10 different orders of presenting the meaning options).}
    \label{fig:john-example}
\end{figure}

Figurative uses of language --- where a speaker does not literally say what they mean --- are ubiquitous in human communication. For example, when faced with a blizzard with heavy winds, a speaker may say ``It's a little chilly out.'' to indicate that it is very cold outside. Similarly, as illustrated in Fig.~\ref{fig:john-example}, a teacher might refer to a student as ``really sharp'' when they intend to convey that they are in fact not very clever. As a result, it is imperative for language technologies and computational models of language to account for this linguistic phenomenon.

However, recent studies have shown that large language models (LLMs) struggle with figurative interpretations of language. For instance, \citet{hu-etal-2023-fine} show that LLMs often misinterpret utterances which subvert listener expectations such as in irony and humour. In Fig.~\ref{fig:john-example}, we show an example where the Mistral-7B-Instruct (V3) LLM overwhelmingly favours an incorrect literal interpretation of the utterance (The student is smart.) over the correct non-literal one (The student is not very clever.). Consequently, this behaviour suggests that LLMs do not correctly or fully model the underlying \emph{communicative goal} (i.e., intention) of utterances needed to interpret them, to the extent that this can be shown through behavioural analysis.

A potential solution to the poor performance of LLMs on figurative understanding is the Rational Speech Act (RSA) framework from probabilistic pragmatics in which the modeling of communicative intents is central. In RSA, an utterance's meaning is interpreted probabilistically by a \emph{pragmatic listener} which reasons about a posited \emph{pragmatic speaker}'s likelihood of generating the utterance under different intended meanings. However, while RSA was designed to interpret utterances based on their communicative goal, its original formulation does not allow non-literal meaning interpretations of utterances. Previous attempts to address this limitation require modeling a speaker's motivations for being non-literal (e.g., to express their affect of annoyance or joy) in a setting-specific way \citep{kao_number_words_2014,Kao2015Irony}.

The key insight that we leverage is that non-literal language usage follows systematic patterns, which can be grouped into \emph{rhetorical strategies} \citep{bertin2016linguistic}. For example, irony can be characterized by an intended meaning being the opposite of its literal meaning, whereas hyperbole involves overstating the intended meaning. Thus, we argue that pragmatic models should reify the rhetorical strategy as a mediating mechanism between language form and communicative intent. To this end, we introduce a novel formulation of the RSA framework: \textbf{R}hetorical-\textbf{S}trategy-\textbf{A}ware \textbf{RSA} \rsatwo. In \rsatwo, a speaker's rhetorical strategy is explicitly modeled as a latent variable and used by the pragmatic listener to interpret an utterance's potentially non-literal intended meaning. 

We show experimentally that \rsatwospace enables pragmatic listeners to infer non-literal interpretations of utterances without needing to model a speaker’s motivations for being non-literal. Using datasets of non-literal number expressions \citep{kao_number_words_2014} and of ironic weather utterances \citep{Kao2015Irony}, we show that \rsatwospace provides non-literal interpretations of utterances which closely match those of humans and which often outperform those made by existing \emph{affect-aware RSA} methods. In addition, we couple \rsatwospace with LLMs and prompt engineering to achieve state-of-the-art performance on the ironic split of an expanded utterance interpretation dataset, PragMega+, which we design for this study. This latter result suggests that probabilistic pragmatics methods may help mitigate LLMs' bias towards literal interpretations and lexical overlap as exhibited in Fig.~\ref{fig:john-example}.

To summarize, in this work, we design \rsatwo, a communicative-goal-centered procedure of figurative language interpretation. By explicitly considering the rhetorical strategies a speaker might use, this framework provides pragmatic non-literal interpretations of utterances which are aligned with human interpretations. In addition, we show that \rsatwospace can be used as a tool to help align LLM interpretations of figurative utterances with their intended meaning. Beyond figurative language understanding, our work re-establishes the importance of computational models of language which focus on modeling the communicative goals of speakers and listeners.  

\section{Related Work}

\label{sec:rw}
\label{sec:human_irony}

\paragraph{Modeling Human Uses of Figurative Language} There has been extensive work to explain why humans use figurative language. Early studies in pragmatics \citep{grice1975logic,horn1984towards} attempt to explain such uses of language by positing that language users communicate with each other ``cooperatively'' by following certain ``conversational maxims''. In this account, figurative statements can result from reconciling violations of conversational maxims with the cooperative principle. Subsequent studies have shown that speakers use figurative expressions to soften the tone of a critique or to be humorous \citep{robertsWhyPeopleUse1994,colstonSaltingWoundSugaring1997}, that figurative uses of language can sometimes be easier to process cognitively than their literal counterparts \citep{ContextualEffectsUnderstanding1979} and that listeners understand figurative statements by explicitly interpreting the mental states of speakers through the ``Theory of Mind''-network part of the brain \citep{spotornoNeuralEvidenceThat2012}.%

\citet{kao_number_words_2014,kao2014formalizing,Kao2015Irony} adapt the RSA framework originally introduced by \citet{frank_ref_game_2012} to enable non-literal interpretations of figurative utterances (e.g., hyperbolic utterances about prices, ironic utterances about the weather). To enable non-literal interpretations, the authors assume that figurative language use is motivated by affect (e.g., to convey joy, annoyance, etc.) and perform joint inference over both intended meaning and affect to achieve non-literal interpretations. In this work, we seek to computationally model figurative language \emph{without} having to explicitly model its motivation which may be difficult to determine in general. For instance, when John ironically says ``This one is really sharp.'' (Fig.~\ref{fig:john-example}), he may be motivated by one (or more) of several reasons: expressing some kind of affect (e.g., humour, frustration), gauging the principal's attention, using a conventionalized way of referring to students, etc. In contrast, \rsatwospace lifts its dependence on explicitly modeling a speaker's motivation to use figurative language and instead directly accounts for the possible rhetorical strategies being used to produce non-literal interpretations of utterances.%

\paragraph{Figurative Language Modeling in NLP} In NLP, the ability of language systems to understand figurative language has mainly been evaluated through detection and generation \citep{li-sporleder-2010-using,tsvetkov-etal-2014-metaphor,van-hee-etal-2018-semeval,balestrucci-etal-2024-im,lai-nissim-2022-multi,lai-etal-2023-multilingual}. A common approach to solving these tasks has been to collect figurative-language-specific datasets and to fine-tune language classification and generation models on them \citep{van-hee-etal-2018-semeval}. With the shift towards LLMs, there has been growing interest in evaluating pre-trained and instruction-tuned LLMs on their ability to recognize, interpret and generate figurative language such as irony and sarcasm \citep{gu-etal-2022-just,liu-etal-2022-testing,hu-etal-2023-fine}. LLMs have been shown to perform significantly worse than humans in interpreting figurative language, such as humour and irony. In this work, we propose a computational model of figurative language that explicitly models rhetorical strategies and which we show can be coupled with LLMs to better align their interpretations with speaker intentions.%

\paragraph{RSA in NLP} \citet{andreas_reasoning_2016} were the first to use neural listeners and speakers with RSA in the context of image captioning. \citet{friedUnifiedPragmaticModels2018} extended the use of RSA to instruction following and generation. \citet{shenPragmaticallyInformativeText2019} investigated RSA for abstractive summarization and \citet{cohn-gordonLostMachineTranslation2019} applied it to machine translation. In closer connection to our work, \citet{careniniFullyInterpretableMore2024} combined the RSA framework with LLMs for the task of metaphor understanding. \citet{tsvilodub2025nonliteralunderstandingnumberwords} leverage affect-aware RSA to enable LLMs to provide non-literal meaning interpretations of utterances with number words. In contrast, we develop a novel RSA framework, \rsatwo, and apply it to several non-literal utterance interpretation tasks.%

\section{Rhetorical-Strategy-Aware RSA}

\label{sec:rsa_two}

In this section, we introduce our novel formulation of the RSA framework --- \textbf{R}hetorical-\textbf{S}trategy-\textbf{A}ware \textbf{RSA} \rsatwo --- which explicitly incorporates the rhetorical strategy into the RSA framework. We first review the inner workings of the RSA framework and then present the novel \rsatwospace framework. Throughout this presentation, we will use the running example of a speaker ironically saying ``The weather is amazing.'' during a blizzard, inspired by \citet{Kao2015Irony}.

\subsection{The RSA Framework}

\label{sec:standard_rsa}

The RSA framework was introduced to model human communication by accounting for listener and speaker expectations, originally in the context of scalar implicature \citep{frank_ref_game_2012}. The goal of this framework is to derive a probability distribution over some fixed (finite) set of meanings $\mathcal{M}$ for some observed utterance $u \in \mathcal{U}$ generated in some context\footnote{The context $c$ can be linguistic (e.g., conversation history) as well as situational (e.g., the current state of the world).} $c \in \mathcal{C}$. To do so, this framework posits the existence of multiple speakers and listeners who reason about each other in a recursive fashion. This recursive procedure begins with a \emph{literal listener} $L_0$, which reasons literally about the interpretation of utterances. Its conditional distribution over $\mathcal{M}$ is defined as
\begin{align}
    P_{L_0}(m|c,u) &\propto \mathds{1}_{m \in \llbracket u \rrbracket} \cdot P_{{M}|{C}}(m|c),
\end{align}
where $P_{{M}|{C}}$ is a prior conditional distribution over all meanings in $\mathcal{M}$ given a context $c$; and $\llbracket \cdot \rrbracket : \mathcal{U} \rightarrow \mathcal{P(M)}$ denotes a shared semantic understanding function\footnote{$\mathcal{P()}$ is the power set function.} which, given some utterance $u \in \mathcal{U}$, returns the set of possible intended meanings which are \emph{literally} compatible with $u$.

For instance, given our running example ``The weather is amazing.'' and a discrete set of weather states which a speaker might want to convey, $\mathcal{M} = \{\texttt{terrible}, \texttt{bad}, \texttt{ok}, \texttt{good}, \texttt{amazing}\}$, the semantic understanding function would return the following set:%
\[
\llbracket \text{``The weather is amazing.''} \rrbracket = \{\texttt{amazing}\}.
\]%
With the base case of the reasoning procedure defined, the RSA framework provides a speaker-aware interpretation of the utterance $u$ by positing the existence of a \emph{pragmatic speaker}, $S_1$, which selects an utterance $u$ based on its prior probability, $P_{{U}|C}$, and $L_0$'s expected information gain.\footnote{While Equation~\ref{eq:rsa_pragmatic_speaker} may appear different from its common expected-utility formulation \citep{goodman2013knowledge}, we show the equivalence of these two formulations in Appendix~\ref{app:equivalence_rsa_formulations}.} That is,
\begin{align}
    \label{eq:rsa_pragmatic_speaker}
    P_{S_1}(u|c,m) &\propto P_{L_0}(m|c,u)^\alpha P_{{U}|{C}}(u|c), 
\end{align}%
where $\alpha$ is a rationality parameter that controls how much the pragmatic speaker will favor selecting the most informative utterance according to the literal listener. Note that the pragmatic speaker is normalized across a set of utterances which are either predefined or generated based on $c$ \citep{andreas_reasoning_2016}.

According to Equation~\ref{eq:rsa_pragmatic_speaker}, a speaker intending to convey meaning $m$ will distribute its probability mass proportional to the probability that $L_0$ infers $m$ from $u$. In our running example, assuming a discrete set of utterances of the form $\mathcal{U} = \{\text{``The weather is $m$.''} : m \in \mathcal{M}\}$, a speaker wishing to convey $m = \texttt{amazing}$ will place \emph{all} of its probability mass on the utterance ``The weather is amazing.'' since no other utterance enables $L_0$ to recover this intended meaning.

Finally, to interpret a given utterance $u$ pragmatically, the \emph{pragmatic listener}, $L_1$, updates their prior belief about the distribution of possible intended meanings $P_{{M}|{C}}$ by using the likelihood that the utterance was generated by $S_1$ in the given context. In other words, the pragmatic listener's conditional probability distribution over $\mathcal{M}$ becomes the following posterior Bayesian update:
\begin{align}
    P_{L_1}(m|c,u) \propto P_{S_1}(u|c,m)  P_{{M}|{C}}(m|c).
\end{align}%

In the case of our ironically uttered statement ``The weather is amazing.'', the pragmatic listener will reason through $S_1$'s generation process and will conclude that they intended to say that the weather is \texttt{amazing} \emph{despite} the blizzard. Thus, the pragmatic listener is unable to escape the literal interpretation of the utterance determined by $L_0$.%

In general, it can be shown that, under the RSA framework, interpretations of utterances which are not compatible with the literal meaning of an utterance will be assigned zero probability by the pragmatic listener (Proof in Appendix~\ref{app:standard_rsa_non_literal}). Existing \emph{affect-aware RSA} approaches discussed in Section~\ref{sec:human_irony} have attempted to mitigate this limitation by expanding the model of the speaker to include a random variable which accounts for their motivation to be figurative (e.g., for affect-related reasons). In contrast, our \rsatwospace approach does not require modeling a speaker's implicit motivations for being non-literal. Rather, the pragmatic listener in \rsatwospace reasons directly and explicitly about the possible rhetorical strategies being employed to achieve non-literal interpretations of utterances.

\subsection{The Rhetorical-Strategy-Aware RSA Framework}

\label{sec:rsatwooriginal}

We introduce a rhetorical-strategy-aware RSA framework, \rsatwo, which defines a rhetorical strategy variable $r \in \mathcal{R}$ and a rhetorical function for each value of $r$, $f_{r}: \mathcal{C} \times \mathcal{M} \times \mathcal{U} \rightarrow [0,1]$. This function generalizes the literal semantic understanding indicator function $\mathds{1}_{m \in \llbracket u \rrbracket}$ to an arbitrary function over $[0,1]$. The rhetorical function enables non-literal interpretations of utterances to arise based on the employed rhetorical strategy. For instance, using our running blizzard example, the ironic and hyperbolic rhetorical strategies might return the following, where $c = \texttt{blizzard}$ is the context variable:%

\noindent 
\resizebox{0.95\linewidth}{!}{%
  \begin{minipage}{\linewidth} 
  \begin{align*}
    f_\textit{irony}(c, \texttt{terrible}, \text{``The weather is amazing.''}) &= 1, \\
    f_\textit{irony}(c, \texttt{amazing}, \text{``The weather is amazing.''}) &= 0, \\
    f_\textit{hyperbole}(c, \texttt{good}, \text{``The weather is amazing.''}) &= 1, \\
    f_\textit{hyperbole}(c, \texttt{amazing}, \text{``The weather is amazing.''}) &= 0.
  \end{align*}
  \end{minipage}
}
\newline\newline
We replace the semantic understanding function with our generalized rhetorical function within the $P_{L_0}$ equation as follows:
\begin{align}
    P_{L_0}(m|c,u,r) &\propto f_r(c,m,u)  P_{M|C}(m|c). \label{eqn:rsatwol0}
\end{align}%
Thus, a meaning which receives zero probability mass when an utterance is interpreted literally by $L_0$ may still receive non-zero probability mass when that same utterance is interpreted using irony, hyperbole or some other non-literal rhetorical strategy. For example, using $f_\textit{irony}$ and $f_\textit{hyperbole}$ in our running example would enable the probability distribution of $L_0$ to shift towards non-literal interpretations of the utterance such as \texttt{terrible} and \texttt{good} respectively. Previous extensions of the RSA framework have modified the semantic understanding indicator function to account for lexical uncertainty \citep{bergen2016pragmatic}. To the best of our knowledge, we are the first to propose this generalization to model figurative language.

We define the pragmatic speaker and listener like in standard RSA, with our rhetorical strategy variable as an additional conditional factor:
\begin{align}%
    P_{S_1}(u|m,c,r) &\propto P_{L_0}(m|c,u,r)^\alpha  P_{U|C}(u|c), \label{eq:rsa_two_pragmatic_speaker} \\
    P_{L_1}(m|c,u,r) &\propto P_{S_1}(u|m,c,r)  P_{M|C}(m|c).
\end{align}%
At inference time, the rhetorical strategy being used is an unobserved latent variable which must be marginalized out while accounting for its probability, $P_{R|CU}(r|c,u)$ on $\mathcal{R}$, as follows:%
\begin{equation}
 P_{L_1}(m|c,u) =   \sum_{r'} P_{L_1}(m|c,u,r') P_{R|CU}(r'|c,u).\label{eq:rsa_marginalization}
\end{equation}%
Returning to our running example, let us consider a discrete set of rhetorical strategies $\mathcal{R} = \{\textit{literal}, \textit{irony}, \textit{hyperbole}\}$. We observe that the pragmatic listener conditioned with the \textit{literal} strategy would produce the same distribution as the standard-RSA-derived pragmatic listener. However, the utterance would also be interpreted by $P_{L_1}(m|c,u,\textit{irony})$ with most probability mass concentrated around \texttt{terrible}, the opposite of the utterance's literal meaning, and by $P_{L_1}(m|c,u,\textit{hyperbole})$ with most mass concentrated around \texttt{good}, the utterance's scaled down literal meaning. By marginalizing these listener distributions using $P_{R|CU}$,\footnote{$P_{R|CU}$ may be uniform, human-derived or computed experimentally.} we would obtain a distribution which assigns mass to meanings beyond the utterance's literal interpretation. In this way, \rsatwospace allows modeling figurative interpretations of language \emph{without} needing to explicitly model the motivations behind its use. In Appendix~\ref{app:qud_rsa_special_case}, we prove that this latter approach to non-literal language is a special case of \rsatwospace wherein the rhetorical strategy function is repurposed to model the motivations of non-literal utterances (e.g., affect).%

\section{Non-Literal Interpretations of Figurative Language with \rsatwo}

\label{sec:study1}

In this first experimental study, we aim to show that \rsatwospace can be used to derive non-literal meaning interpretations which match human interpretations as well as by existing affect-aware RSA methods. We implement \rsatwospace in two settings: non-literal number price expressions (e.g., ``This kettle costs 10000\$.'') and ironic weather utterances (e.g., ``The weather is amazing.'' during a winter blizzard). We show that \rsatwospace produces utterance interpretations that are on par with or better than affect-aware RSA interpretations.%

\subsection{Datasets}

\paragraph{Non-literal number expressions.} We use the dataset from \citet{kao_number_words_2014} which contains a collection of literal and non-literal number expressions related to the price of objects. The context space $\mathcal{C}$ is the set of objects being described, the meaning space $\mathcal{M}$ is the price of the object being described and the utterance space $\mathcal{U}$ is a verbalization of the object's price. These three sets are listed below:%
\begin{align*}
    \mathcal{C} &= \{\text{electric kettle}, \text{laptop}, \text{watch}\}, \\
    \mathcal{M} &= \{50, 51, 500, 501, 1000, 1001, \\ 
    & \quad 5000, 5001, 10000, 10001\}, \\ 
    \mathcal{U} &= \{\text{``The $c$ costs $m$ dollars.''} : c \in \mathcal{C}, m \in \mathcal{M}\}.
\end{align*}%
The authors collected meaning and affect priors $P(m|c), P(a|c)$ from 30 human participants as well as meaning and affect posteriors $P(m|c,u), P(a|c,u)$ from 120 human participants via a Likert-scale probability elicitation technique.%

To enable \rsatwo-based non-literal interpretations, we define the space of rhetorical strategies $\mathcal{R}$ as consisting of four types: \emph{literal} (describing the price exactly), \emph{hyperbole} (overstating the price), \emph{understatement} (understating the price) and \emph{halo} ( providing a round figure rather than the exact one).%

\paragraph{Ironic weather utterances.} We use the dataset from \citet{Kao2015Irony} which contains a collection of utterances about the weather similar to the one from our running example in Section~\ref{sec:rsa_two}. In this case, the context space $\mathcal{C}$ is represented visually through images depicting different weather conditions (see Appendix~\ref{app:weather_images} for the images), the utterance space $\mathcal{U}$ consists of statements about the weather (e.g., ``The weather is amazing.''), and the meaning space $\mathcal{U}$ corresponds to the true weather state being communicated by the speaker. These three sets are listed below:%
\begin{align*}
    \mathcal{C} &= \{c_i : i \in [1 \dots 9]\}, \\
    \mathcal{M} &= \{\texttt{terrible},\texttt{bad},\texttt{ok},\texttt{good},\texttt{amazing}\}, \\
    \mathcal{U} &= \{\text{``The weather is $m$''} : m \in \mathcal{M}\}.
\end{align*}%
The context space contains three types of contexts: those where the weather is visibly good ($\{c_1,c_2,c_3\}$), those where the weather is visibly bad ($\{c_7,c_8,c_9\}$) and those where the weather is neither visibly good nor bad ($\{c_4,c_5,c_6\}$). The original authors of the paper collected meaning and affect priors $P(m|c), P(a|c)$ from 49 human participants as well as meaning, affect and irony posteriors $P(m|c,u), P(a|c,u), P(r|c,u)$ from 120 human participants using both normalized counts and a Likert-scale probability elicitation technique similar to \citet{kao_number_words_2014}.%

To enable \rsatwo-based non-literal interpretations, the space of rhetorical strategies $\mathcal{R}$ is defined using both the \textit{literal} (describing the weather as it is perceived) and \textit{irony} (describing the weather as the opposite of how it is perceived) rhetorical strategies.

\subsection{RSA Model Experiments}

\subsubsection{Baseline}

\paragraph{Affect-aware RSA.} For the non-literal number expressions dataset, we use the pragmatic listener as computed by \citet{kao_number_words_2014}. For the weather utterance dataset, we re-implement both the affect-aware literal and pragmatic listeners (see Appendix~\ref{app:reimplementation_details} for implementation details).

\subsubsection{\rsatwospace Implementations}

\paragraph{Non-literal number expressions.} We manually define $f_r$ based on the intuitive definitions of each of the rhetorical strategies in $\mathcal{R}$. To simplify notation, we only consider the integer parts of the utterance i.e., $u = \text{``The $c$ costs $y$ dollars.''}$ becomes $u = y$. We define $f_r$ as:%

\noindent 
\resizebox{0.8\linewidth}{!}{%
  \begin{minipage}{\linewidth} 
  \begin{align*}
  f_r(c, m, u) =
  \begin{cases}
      1       & \text{if } r = \text{literal}, u = m \\
      1       & \text{if } r = \text{hyperbole}, \\
              & \quad u - m > 10 \\ 
      1       & \text{if } r = \text{understatement}, \\
              & \quad m - u > 10 \\
      1       & \text{if } r = \text{halo}, |u - m| = 1, \\
              & \quad u \text{ ends in 0} \\
      0.001   & \text{otherwise}
  \end{cases}
  \end{align*}
  \end{minipage}
}
\newline\newline
Thus, for instance, if a kettle's price is $m = 50$, but the utterance used is $u = 500$, then this will trigger the \emph{hyperbole} rhetorical strategy. Similarly, if the kettle's price is $m = 501$ and the utterance is $u = 500$, then this will activate the \emph{halo} rhetorical strategy. We ignore the context $c$ in $f_r$ for simplicity, although in principle the effect of a rhetorical strategy like hyperbole could vary across objects. We substitute $f_r$ in $L_0$'s Equation~\ref{eqn:rsatwol0} and use the meaning prior $P(m|c)$ derived from \citet{kao_number_words_2014} and a uniform rhetorical strategy prior, $P(r|c,u) = \frac{1}{4}$, for the marginalization of Equation~\ref{eq:rsa_marginalization}. We set the rationality parameter $\alpha = 1$.

\paragraph{Ironic weather utterances.} To highlight the flexibility of \rsatwo, we approximate the rhetorical function $f_r$ across all 450 $(c,m,u,r)$ quadruples\footnote{$9\text{ contexts }\times5\text{ meanings }\times5\text{ utterances }\times2\text{ rhetorical strategies}$} using a neural network. Specifically, we split the dataset into training, validation and test sets ($60\%/20\%/20\%$) and train a 2-layer neural network ($16 \times 16 \times 5$) with sigmoid activations. We encode the contexts, meanings, utterances, and rhetorical strategies using a one-hot encoding and use a cross-entropy loss between $P_{L_1}(m|c,u)$ and the meaning selected by human participants given the context $c$ and the utterance $u$. The rationality parameter is set to $\alpha = 1$. We use the human-elicited probability $P(r|c,u)$ for the marginalization across the two rhetorical strategies in Equation~\ref{eq:rsa_marginalization}. Additional details can be found in Appendix~\ref{app:ironic_weather_utterances_exp_details}.

\subsection{Results}

\begin{figure}[h]
    \centering
    \includegraphics[width=\linewidth]{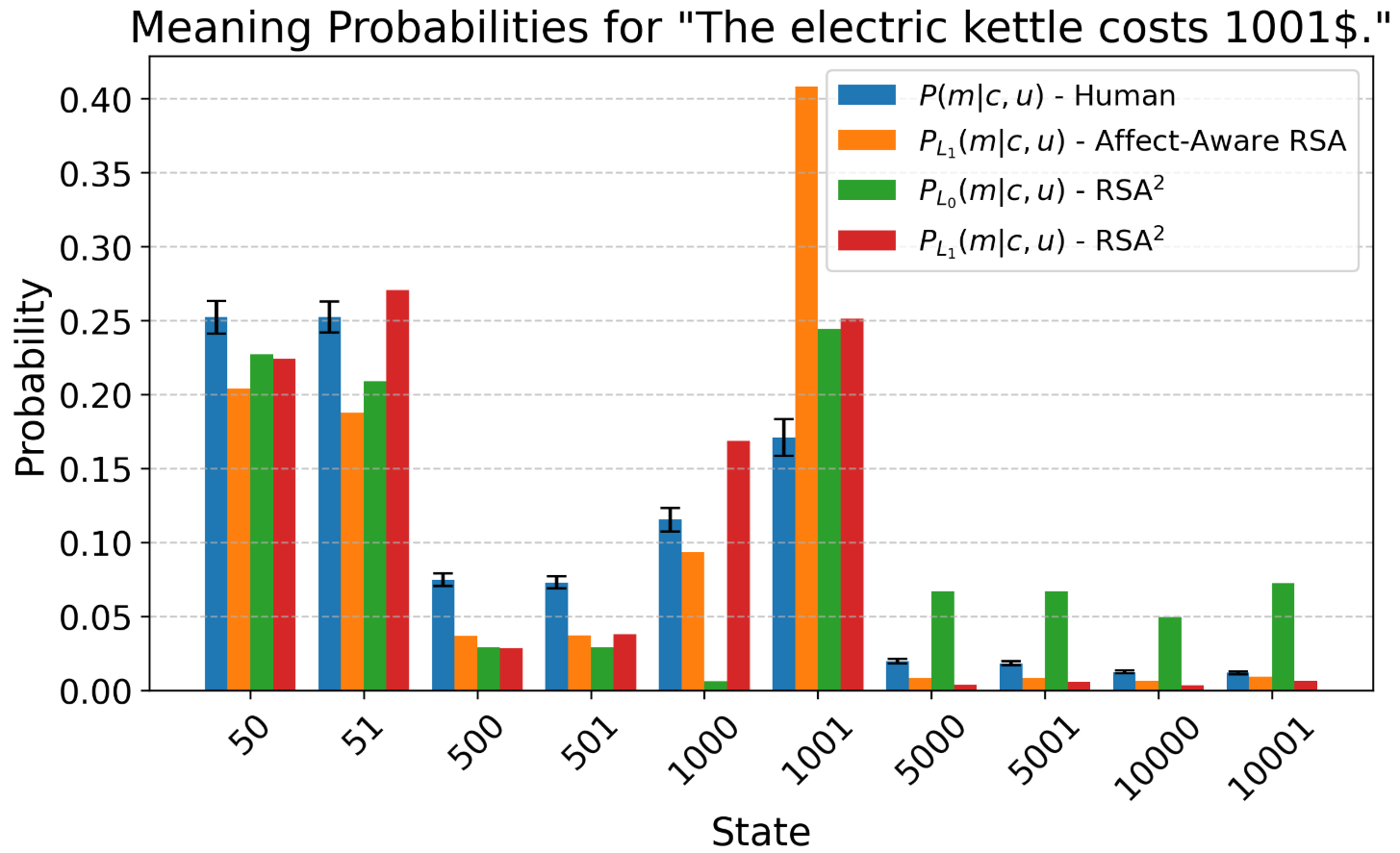}
    \caption{Meaning distribution of human posteriors and of affect-aware and \rsatwospace listeners for the utterance ``The electric kettle costs 1001 dollars''.}
    \label{fig:the_kettle}
\end{figure}

\begin{figure}[h]
    \centering
    \includegraphics[width=\linewidth]{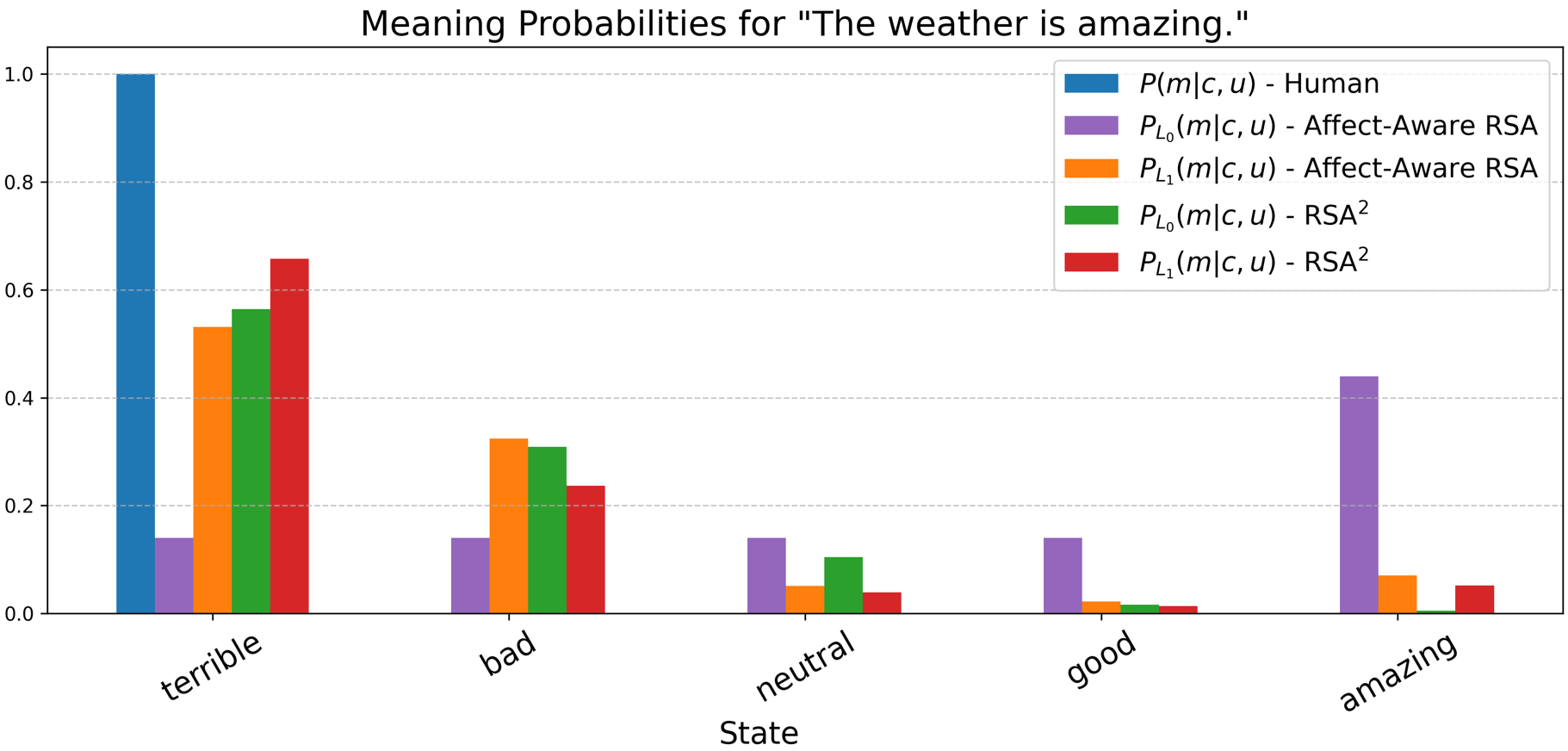}
    \caption{Meaning distribution of human posteriors and of affect-aware and \rsatwospace listeners for the utterance ``The weather is amazing.'' in the context of a blizzard (Image $c_8$ in Appendix~\ref{app:weather_images}).}
    \label{fig:the_weather_is_amazing}
\end{figure}

We plot the meaning distribution of human posteriors and of affect-aware and \rsatwospace listeners for the hyperbolic number expression ``The electric kettle costs 1001 dollars.'' in Fig.~\ref{fig:the_kettle} and for the ironic expression ``The weather is amazing.'' uttered in the context of a blizzard in Fig.~\ref{fig:the_weather_is_amazing}. In both cases, we observe that the \rsatwospace listeners, especially the pragmatic listeners, induce human-like meaning distributions. Figures for all other context–meaning–utterance triples can be found in Appendices~\ref{app:non_literal_number_expressions} and~\ref{app:ironic_weather_utterances_add_results}.

We evaluate our model's predictive power by computing the mean absolute difference (MAD) between the human-derived meaning probability distributions and those generated by the listeners in both affect-aware RSA and \rsatwospace (Table~\ref{tab:mad_numbers}). We note that \rsatwospace is competitive with affect-aware RSA, surpassing it on the ironic weather utterances dataset. While the affect-aware listener outperforms our \rsatwospace listeners on non-literal number expressions, we believe this is due to the poorly calibrated uniform rhetorical strategy posterior we use, which sometimes causes probability mass to be distributed to the tail ends of the meaning space. For instance, while the expression ``The kettle costs 51 dollars.'' is most likely to be uttered with the \textit{literal} rhetorical strategy, a significant portion of the meaning mass is distributed across the rest of meaning space due to the non-zero posterior probability of the \textit{understatement} rhetorical strategy. Overall, we believe this initial study demonstrates that \rsatwospace can induce listener meaning distributions for figurative language that are at least as compatible with human interpretations as those produced by existing affect-aware RSA models.

\begin{table}[h]
\centering
\resizebox{\linewidth}{!}{
    \begin{tabular}{clcc}
        \toprule
        Model & $L_i$ & Non-literal Numbers \(\downarrow\) & Weather Utterances \(\downarrow\) \\
        \midrule
        \multirowcell{2}{Affect-Aware \\ RSA} & $L_0$ & - & 0.2377 \\
        & $L_1$ & \textbf{0.0436} & 0.1278 \\
        \cmidrule{2-4}
        \multirowcell{2}{\rsatwo} & $L_0$ & 0.0438 & 0.1647 \\
        & $L_1$ & 0.0467 & \textbf{0.1229} \\
        \bottomrule
    \end{tabular}
}
\caption{Mean absolute differences between listener meaning distributions, $P_{L_i}(m|c,u), i = 0,1$, and the human posterior, $P(m|c,u)$, for both affect-aware and \rsatwospace on both the non-literal number expressions and ironic weather utterances datasets.}
\label{tab:mad_numbers}
\end{table}

\section{LLM Irony Interpretation with \rsatwo}

\label{sec:experiment2}

In this section, we apply \rsatwospace to LLMs to generate non-literal interpretations of utterances that better align with human interpretations of figurative language. In particular, we focus on the task of interpreting ironic utterances in situational contexts beyond the weather examples from Section~\ref{sec:study1}. Our results demonstrate that integrating LLMs into \rsatwospace can, to a certain extent, mitigate their biases toward lexical-overlap and literal interpretations.

\subsection{The PragMega+ Dataset}

\label{sec:dataset}

We expand the PragMega dataset from \citet{hu-etal-2023-fine} to evaluate the \rsatwospace framework on ironic utterance interpretation. The original PragMega dataset contains 25 ironic scenarios where the intended meaning of an utterance is non-literal, such as in Fig.~\ref{fig:john-example}. Each scenario includes a background context $c$ and an utterance $u$ produced ironically by a speaker. In addition, each scenario is accompanied by four candidate intended meanings, which act as our meaning space$\mathcal{M}$: the literal meaning of $u$ (\texttt{Literal Meaning}), the intended meaning of $u$ (\texttt{Non-Literal Meaning}), a lexical overlap distractor meaning (\texttt{Overlap Meaning}), and a non-sequitur distractor meaning (\texttt{Non-Sequitur Meaning}).

Our expanded dataset, PragMega+, adds 25 ironic scenarios in the same format as part of our test set. In addition, we manually modify each of the 50 scenarios to create 50 \emph{literal} scenarios where the intended meaning is the \texttt{Literal Meaning}. Examples of scenarios with \texttt{Non-Literal} and \texttt{Literal} intended meanings from PragMega+ can be found in Appendix~\ref{app:pragmega_more_examples}.

\subsection{Experimental Setup}

\label{sec:experimental_setup}

\subsubsection{LLM Probability Estimation}

\label{sec:llms}

To integrate LLMs within the \rsatwospace framework, we use an LLM $N$ to estimate all conditional probabilities of the form $P_{N}(y|x)$. To do so, we use a prompt template similar to \citet{hu-etal-2023-fine} which, given $x$, lists all possible values of $y$ in a multiple-choice question (MCQ) format. After passing the prompt to the LLM $N$, we extract the next-token logit of the number corresponding to each option. These logits are re-normalized so that the sum of their corresponding probabilities equals one. In addition, to avoid positional bias, we randomly shuffle the order of the options in the prompt $10$ times\footnote{If the total number of permutations is less than $10$, we use that total instead.} and average across shuffles to obtain $P_{N}(y|x)$. We use this approach to estimate the meaning prior $P_{N}(m|c)$, the rhetorical strategy posterior $P_{N}(r|c,u)$, the meaning posterior $P_{N}(m|c,u)$, and the meaning posterior \emph{conditioned on} the rhetorical strategies, $P_{N}(m|c,u,r)$, where $\mathcal{R} = \{\textit{literal},\textit{irony}\}$. We experiment with two instruction-tuned models: Mistral-7B-Instruct (V3) LLM \citep{jiang2023mistral7b} and Llama-8B-Instruct (V3.1) \citet{llama3modelcard} via HuggingFace \citep{wolf2020huggingfacestransformersstateoftheartnatural}. We performed prompt engineering on the validation set. All prompts used on the test set are presented in Appendix~\ref{app:prompts}.

\subsubsection{Alternative Utterance Generation}

\label{sec:alt_utt_gen}

We use an LLM $G$ conditioned on a scenario's context $c$ to generate the alternative utterances needed to compute the pragmatic speaker's distribution. To do so, we condition $G$'s next-token generation on the full PragMega+ scenarios up to the original utterance's first quotation mark (e.g., \emph{... John said,``} in Fig.~\ref{fig:john-example}), and continue generation until another quotation mark is produced. We use a decoding temperature of $1.0$ and generate $50$ alternative utterances for each context. We also use the LLM’s generation likelihood, $P_G(u \mid c)$, as the utterance prior in our model. We experiment with the base versions of Mistral-7B (V3) \citep{jiang2023mistral7b} and Llama-8B (V3.1) \citep{llama3modelcard}.

\subsubsection{LLM Listeners}

\paragraph{LLM RSA} We implement an LLM RSA baseline by setting $P_{L_0}(m|c,u)$ equal to the raw LLM-derived probabilities, $P_{N}(m|c,u)$. This baseline follows prior neural listener work \citep{andreas_reasoning_2016, monroe-etal-2017-colors} which approximate $P_{L_0}(m|c,u)$ using trained neural networks. We set $\alpha = 1$.


\paragraph{LLM \rsatwo} We implement our LLM \rsatwospace listeners by setting $P_{L_0}(m|c,u,r)$ equal to the LLM-derived probabilities $P_{N}(m|c,u,r)$. For the rhetorical strategy posterior we use both the raw LLM-derived probabilities, $P_{N}(r|c,u)$, and an indicator posterior, $I(r|c,u) = \mathds{1}_{r = \arg\max_{r'} P_N(r'|c,u)}$. We set $\alpha = 1$.

\paragraph{Where is $f_r(c,m,u)$ in LLM \rsatwo?} Based on the LLM \rsatwospace literal listener, the rhetorical strategy function is implicitly set as:%
\begin{equation}
    f_r(c,m,u) = \frac{P_{N}(m|c,u,r)}{k \cdot P_{N}(m|c)},\label{eq:what_is_frcmu}
\end{equation}%
where $k = \max_{m'} f_r(c,m',u)$ ensures that $f_r(c,m,u) \in [0,1]$.

\paragraph{Why is affect-aware RSA not included as a baseline?} While it is in principle possible to implement an LLM-based Affect-Aware RSA baseline \citep{tsvilodub2025nonliteralunderstandingnumberwords}, we do not pursue this direction, as it would require defining a suitable affect space for each situational context.

\subsection{Results}

Table~\ref{tab:meaning_scores} presents the average listener meaning probabilities for the correct, incorrect and distractor meanings (aggregated) across all 50 scenarios in the test set. We use the LLM pair $G = \text{Llama-8B (V3.1)}$ and $N = \text{Mistral-7B-Instruct (V3)}$, since this pair performed best on the validation set. Listener probabilities conditioned on each rhetorical strategy are shown in Table~\ref{tab:p_r_c_u}. Overall, we see that the LLM \rsatwospace listeners outperform the baseline LLM RSA listeners. In particular, the LLM \rsatwospace literal listener marginalized using the indicator rhetorical strategy posterior, $I(r|c,u)$, performs the best while the baseline LLM RSA literal listener performs the worst.


\begin{table}[h]
    \centering
    \resizebox{\linewidth}{!}{
    \begin{tabular}{llccc}

    \toprule
    Model & $L_i$ & Correct \(\uparrow\) & Incorrect \(\downarrow\) & Distractor \(\downarrow\) \\
    \midrule
    LLM RSA & $L_0$ & 0.73 & 0.24 & 0.02 \\
    & $L_1$ & 0.76 & 0.22 & 0.01 \\
    \cmidrule{2-5}
    \makecell{LLM \rsatwo\\with $P_N(r|c,u)$} & $L_0$ & 0.74 & 0.23 & 0.02 \\
    & $L_1$ & 0.80 & 0.16 & 0.02 \\
    \cmidrule{2-5}
    \makecell{LLM \rsatwo\\with $I(r|c,u)$} & $L_0$ & \textbf{0.85} & \textbf{0.13} & 0.01 \\
    & $L_1$ & 0.84 & 0.13 & \textbf{0.01} \\
    \bottomrule

    \end{tabular}
    }
    \caption{Average listener probabilities, $P_{L_i}(m|c,u), i = 0,1$ for the correct, incorrect and distractor intended meanings on the test set averaged across all 50 scenarios.}
    \label{tab:meaning_scores}
\end{table}

\begin{figure*}[ht]
    \centering
    \includegraphics[width=\linewidth]{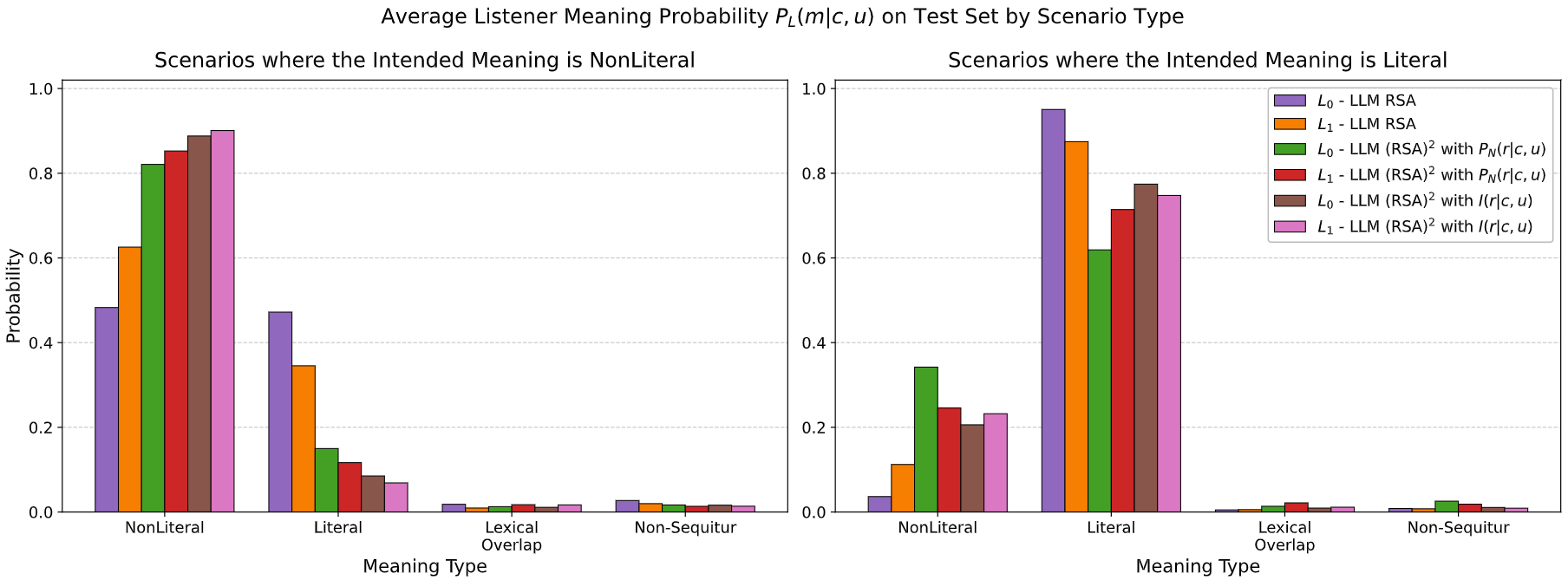}
    \caption{Average listener probabilities on the test set, split between the first 25 scenarios where the intended meaning is non-literal and the last 25 scenarios where the intended meaning is literal.}
    \label{fig:pragmega_nonliteral_literal}
\end{figure*}

We study these results more carefully by plotting the listener probability distributions split between ironic and literal scenarios in Fig.~\ref{fig:pragmega_nonliteral_literal}. On ironic scenarios, the LLM \rsatwospace listener probabilities for the correct non-literal interpretation are all above $0.8$. In contrast, the LLM RSA literal listener assigns a probability of less than $0.5$ ($0.48$) to the correct non-literal intended meaning. However, on literal scenarios, the trend is reversed. The LLM RSA literal listener assigns most of its probability mass ($0.95$) to the correct literal meaning while the best performing of the LLM \rsatwospace listeners -- $L_0$ marginalized with $I(r|c,u)$ -- assigns $0.77$ of its mass to the correct literal intended meaning.

While the strong performance of the LLM RSA listeners on the literal scenarios is expected -- the correct intended meaning is consistent with the utterance and the context -- the drop in LLM \rsatwospace performance in these cases is surprising. Further analysis reveals that this result stems from the listener marginalization, specifically the rhetorical strategy posterior $P_N(r \mid c, u)$. This posterior performs poorly in the literal scenarios where the utterance is intended to be interpreted literally (Table~\ref{tab:p_r_c_u}). While for ironic scenarios the average probability of $P_N(r=\textit{irony}|c,u) = 0.88$ is reasonable, it is far worse on the sincere scenarios with $P_N(r=\textit{literal}|c,u) = 0.55$. This asymmetry in rhetorical strategy posteriors also explains why using $I(r|c,u)$ in the marginalization helps the LLM \rsatwospace listeners. The indicator function blocks more of the probability mass associated with the incorrect meaning from being incorporated into the marginalization.

\subsection{Probability Ablation Study}

\label{sec:analysis_pragmega_plus}

To study the impact of the RSA reasoning process on LLMs, we run an ablation study in which we set the probabilities $P(m \mid c)$ and $P(u \mid c)$ to uniform in both LLM RSA and LLM \rsatwo. Our analysis reveals that ablating $P(u|c)$ does not hinder performance and may slightly improve it. Surprisingly, ablating $P(m|c)$ reveals that most of the performance gain obtained by $L_1$ comes from the meaning prior \emph{and not} from the RSA reasoning process itself. We attribute this result to the way in which we generate alternative utterances (Section~\ref{sec:alt_utt_gen}) which tend to produce literal paraphrases of the intended meaning. For instance, Llama-8B (V3.1) generates alternatives to ``This one is really sharp.'' from Fig.~\ref{fig:john-example} such as ``Her scores were well below average.'' Thus, the pragmatic speaker conditioned on the correct intended meaning distributes probability mass across many compatible alternative utterances, leading to a Bayesian update that \emph{spreads}, rather than \emph{narrows}, the distribution over meanings. Future work on coupling LLMs with RSA may want to ensure diversity in the alternatives generated, for instance by conditioning the generation on individual meanings rather than the context.

\begin{table}[h]
\centering
    \resizebox{\linewidth}{!}{
    \begin{tabular}{llcc} 
    \toprule
    Model & $L_i$ & w/o $P(m|c)$ & w/o $P(u|c)$ \\
    \midrule
    LLM RSA & $L_1$ & 0.44 (-42.7\%) & 0.78 (+1.8\%) \\
    \cmidrule{2-4}
    \makecell{LLM \rsatwo\\with $P_N(r|c,u)$} & $L_1$ & 0.44 (-44.8\%) & 0.80 (+0.3\%) \\
    \cmidrule{2-4}
    \makecell{LLM \rsatwo\\with $I(r|c,u)$} & $L_1$ & 0.51 (-39.4\%) & 0.84 (+0.2\%) \\
    \bottomrule
    \end{tabular}
    }
    \caption{Pragmatic listener probability ablations. We report the average listener posterior probabilities of the correct meaning on the test set (across all 50 scenarios) and the relative change with respect to the unablated model.}
    \label{tab:probability_ablations}
\end{table}

Overall, despite RSA reasoning offering limited improvements, the results obtained with $L_0$ still suggest that explicitly modeling rhetorical strategies may help mitigate lexical-overlap and literal interpretation biases in LLMs.

\section{Conclusion}

In conclusion, we presented \rsatwo, a computational model of figurative language based on the RSA framework that explicitly incorporates rhetorical strategy to support non-literal interpretation. We show that \rsatwospace enables human-compatible interpretations of figurative utterances—including non-literal number expressions and ironic weather statements—\emph{without} explicitly modeling the speaker’s motivations behind using figurative language. We further demonstrated that combining \rsatwospace with LLMs yields state-of-the-art performance on the ironic split of the PragMega+ dataset. We hope this work inspires future research to incorporate pragmatic reasoning techniques within language technologies.

\section*{Limitations}

\paragraph{Limited dataset size and diversity.} While our experimental and theoretical results (Appendix~\ref{app:qud_rsa_special_case}) support the usefulness and applicability of \rsatwo, we believe that future work in computational pragmatics should include broader examples across languages and cultural contexts. Work such as \citet{park-etal-2024-multiprageval,sravanthi-etal-2024-pub} has begun to address these issues. Regarding \rsatwospace specifically, while we have demonstrated its applicability to irony, hyperbole, understatement, and pragmatic halo, evaluating our framework on additional pragmatic phenomena such as metaphor and politeness are promising directions for future research.

\paragraph{Alternative utterance generation.} As discussed in Section~\ref{sec:analysis_pragmega_plus}, pragmatic reasoning does not yield the typical probability mass narrowing associated with RSA due to the distribution of mass across multiple alternative utterances that effectively ``mean'' the same thing within this framework. We believe that this poses an interesting research problem for the field of computational pragmatics in particular (e.g. How do we generate relevant and non-overlapping alternative utterances?), and for natural language processing (e.g., LLM decoding) more broadly.

\paragraph{What if the rhetorical strategies are not known in advance?} While \rsatwospace addresses the challenge of explicitly modeling affect in affect-aware RSA, it introduces a new question: how to determine which rhetorical strategies are appropriate in a given context? We take a first step toward this in Appendix~\ref{app:rsc_rsa}, where we propose and test a clustering-based algorithm that automatically induces \emph{rhetorical strategy clusters}. While our algorithm underperforms LLM \rsatwo, we believe that developing a more principled approach to inducing rhetorical strategies is an important direction for future work.

\section*{Ethics Statement}

We propose \rsatwospace as a computational model of figurative language that aims to better capture human interpretations of non-literal utterances and to help LLMs interpret such utterances in a way that aligns more closely with their intended meaning. We recognize, however, that figurative language differs significantly across languages and cultures. Our model has not yet been validated on languages beyond English, or on figurative phenomena that may be less common in English. We encourage future work—similar to \citet{park-etal-2024-multiprageval}—to explore figurative language in a range of linguistic and cultural contexts beyond those commonly associated with English.

\section*{Acknowledgments}

The authors would like to thank the reviewers for their valuable comments. We would also like to thank Gaurav Kamath, Arkil Patel and Siva Reddy for their insightful comments on an earlier version of this paper. We are also extremely grateful to Justine Kao and Polina Tsvilodub for sharing the non-literal number expressions and ironic weather utterances datasets and for their helpful feedback. This work was supported by the Fonds de Recherche du Québec – Nature et Technologies (FRQNT) and the Natural Sciences and Engineering Research Council of Canada (NSERC). Jackie Chi Kit Cheung is supported by Canada CIFAR AI Chair program. We acknowledge material support from NVIDIA Corporation in the form of computational resources provided to Mila.

\bibliographystyle{acl_natbib}

\bibliography{custom}

\begin{thebibliography}{37}
\expandafter\ifx\csname natexlab\endcsname\relax\def\natexlab#1{#1}\fi

\bibitem[{AI@Meta(2024)}]{llama3modelcard}
AI@Meta. 2024.
\newblock \href {https://github.com/meta-llama/llama3/blob/main/MODEL_CARD.md} {Llama 3 model card}.

\bibitem[{Andreas and Klein(2016)}]{andreas_reasoning_2016}
Jacob Andreas and Dan Klein. 2016.
\newblock \href {https://doi.org/10.18653/v1/D16-1125} {Reasoning about {Pragmatics} with {Neural} {Listeners} and {Speakers}}.
\newblock In \emph{Proceedings of the 2016 {Conference} on {Empirical} {Methods} in {Natural} {Language} {Processing}}, pages 1173--1182, Austin, Texas. Association for Computational Linguistics.

\bibitem[{Balestrucci et~al.(2024)Balestrucci, Casola, Lo, Basile, and Mazzei}]{balestrucci-etal-2024-im}
Pier~Felice Balestrucci, Silvia Casola, Soda~Marem Lo, Valerio Basile, and Alessandro Mazzei. 2024.
\newblock \href {https://doi.org/10.18653/v1/2024.findings-emnlp.847} {{I}{'}m sure you{'}re a real scholar yourself: Exploring ironic content generation by large language models}.
\newblock In \emph{Findings of the Association for Computational Linguistics: EMNLP 2024}, pages 14480--14494, Miami, Florida, USA. Association for Computational Linguistics.

\bibitem[{Bergen et~al.(2016)Bergen, Levy, and Goodman}]{bergen2016pragmatic}
Leon Bergen, Roger Levy, and Noah Goodman. 2016.
\newblock Pragmatic reasoning through semantic inference.
\newblock \emph{Semantics and Pragmatics}, 9:20--1.

\bibitem[{Bertin et~al.(2016)Bertin, Atanassova, Sugimoto, and Lariviere}]{bertin2016linguistic}
Marc Bertin, Iana Atanassova, Cassidy~R Sugimoto, and Vincent Lariviere. 2016.
\newblock The linguistic patterns and rhetorical structure of citation context: an approach using n-grams.
\newblock \emph{Scientometrics}, 109:1417--1434.

\bibitem[{Buitinck et~al.(2013)Buitinck, Louppe, Blondel, Pedregosa, Mueller, Grisel, Niculae, Prettenhofer, Gramfort, Grobler, Layton, VanderPlas, Joly, Holt, and Varoquaux}]{sklearn_api}
Lars Buitinck, Gilles Louppe, Mathieu Blondel, Fabian Pedregosa, Andreas Mueller, Olivier Grisel, Vlad Niculae, Peter Prettenhofer, Alexandre Gramfort, Jaques Grobler, Robert Layton, Jake VanderPlas, Arnaud Joly, Brian Holt, and Ga{\"{e}}l Varoquaux. 2013.
\newblock {API} design for machine learning software: experiences from the scikit-learn project.
\newblock In \emph{ECML PKDD Workshop: Languages for Data Mining and Machine Learning}, pages 108--122.

\bibitem[{Carenini et~al.(2024)Carenini, Bischetti, Schaeken, and Bambini}]{careniniFullyInterpretableMore2024}
Gaia Carenini, Luca Bischetti, Walter Schaeken, and Valentina Bambini. 2024.
\newblock \href {https://doi.org/10.48550/ARXIV.2404.02983} {Towards a {Fully} {Interpretable} and {More} {Scalable} {RSA} {Model} for {Metaphor} {Understanding}}.
\newblock Publisher: arXiv Version Number: 1.

\bibitem[{Cohn-Gordon and Goodman(2019)}]{cohn-gordonLostMachineTranslation2019}
Reuben Cohn-Gordon and Noah Goodman. 2019.
\newblock \href {http://arxiv.org/abs/1902.09514} {Lost in {Machine} {Translation}: {A} {Method} to {Reduce} {Meaning} {Loss}}.
\newblock ArXiv:1902.09514 [cs].

\bibitem[{Colston(1997)}]{colstonSaltingWoundSugaring1997}
Herbert~L. Colston. 1997.
\newblock \href {https://doi.org/10.1080/01638539709544980} {Salting a wound or sugaring a pill: {The} pragmatic functions of ironic criticism}.
\newblock \emph{Discourse Processes}, 23(1):25--45.
\newblock Publisher: Routledge \_eprint: https://doi.org/10.1080/01638539709544980.

\bibitem[{Frank and Goodman(2012)}]{frank_ref_game_2012}
Michael~C. Frank and Noah~D. Goodman. 2012.
\newblock \href {https://doi.org/10.1126/science.1218633} {Predicting {Pragmatic} {Reasoning} in {Language} {Games}}.
\newblock \emph{Science}, 336(6084):998--998.
\newblock Publisher: American Association for the Advancement of Science.

\bibitem[{Fried et~al.(2018)Fried, Andreas, and Klein}]{friedUnifiedPragmaticModels2018}
Daniel Fried, Jacob Andreas, and Dan Klein. 2018.
\newblock \href {https://doi.org/10.18653/v1/N18-1177} {Unified {Pragmatic} {Models} for {Generating} and {Following} {Instructions}}.
\newblock In \emph{Proceedings of the 2018 {Conference} of the {North} {American} {Chapter} of the {Association} for {Computational} {Linguistics}: {Human} {Language} {Technologies}, {Volume} 1 ({Long} {Papers})}, pages 1951--1963, New Orleans, Louisiana. Association for Computational Linguistics.

\bibitem[{Gibbs~Jr.(1979)}]{ContextualEffectsUnderstanding1979}
Raymond~W. Gibbs~Jr. 1979.
\newblock \href {https://doi.org/10.1080/01638537909544450} {Contextual effects in understanding indirect requests}.
\newblock \emph{Discourse Processes}, 2(1):1--10.
\newblock Publisher: Routledge \_eprint: https://doi.org/10.1080/01638537909544450.

\bibitem[{Goodman and Stuhlmüller(2013)}]{goodman2013knowledge}
Noah~D Goodman and Andreas Stuhlmüller. 2013.
\newblock Knowledge and implicature: Modeling language understanding as social cognition.
\newblock \emph{Topics in cognitive science}, 5(1):173--184.

\bibitem[{Grice(1975)}]{grice1975logic}
HP~Grice. 1975.
\newblock Logic and conversation.
\newblock \emph{Syntax and Semantics}, 3:43--58.

\bibitem[{Gu et~al.(2022)Gu, Fu, Pyatkin, Magnusson, Dalvi~Mishra, and Clark}]{gu-etal-2022-just}
Yuling Gu, Yao Fu, Valentina Pyatkin, Ian Magnusson, Bhavana Dalvi~Mishra, and Peter Clark. 2022.
\newblock \href {https://doi.org/10.18653/v1/2022.flp-1.12} {Just-{DREAM}-about-it: Figurative language understanding with {DREAM}-{FLUTE}}.
\newblock In \emph{Proceedings of the 3rd Workshop on Figurative Language Processing (FLP)}, pages 84--93, Abu Dhabi, United Arab Emirates (Hybrid). Association for Computational Linguistics.

\bibitem[{Horn(1984)}]{horn1984towards}
Laurence Horn. 1984.
\newblock Towards a new taxonomy for pragmatic inference: Q-and r-based implicature.
\newblock \emph{Meaning, form and use in context}.

\bibitem[{Hu et~al.(2023)Hu, Floyd, Jouravlev, Fedorenko, and Gibson}]{hu-etal-2023-fine}
Jennifer Hu, Sammy Floyd, Olessia Jouravlev, Evelina Fedorenko, and Edward Gibson. 2023.
\newblock \href {https://doi.org/10.18653/v1/2023.acl-long.230} {A fine-grained comparison of pragmatic language understanding in humans and language models}.
\newblock In \emph{Proceedings of the 61st Annual Meeting of the Association for Computational Linguistics (Volume 1: Long Papers)}, pages 4194--4213, Toronto, Canada. Association for Computational Linguistics.

\bibitem[{Jiang et~al.(2023)Jiang, Sablayrolles, Mensch, Bamford, Chaplot, de~las Casas, Bressand, Lengyel, Lample, Saulnier, Lavaud, Lachaux, Stock, Scao, Lavril, Wang, Lacroix, and Sayed}]{jiang2023mistral7b}
Albert~Q. Jiang, Alexandre Sablayrolles, Arthur Mensch, Chris Bamford, Devendra~Singh Chaplot, Diego de~las Casas, Florian Bressand, Gianna Lengyel, Guillaume Lample, Lucile Saulnier, Lélio~Renard Lavaud, Marie-Anne Lachaux, Pierre Stock, Teven~Le Scao, Thibaut Lavril, Thomas Wang, Timothée Lacroix, and William~El Sayed. 2023.
\newblock \href {http://arxiv.org/abs/2310.06825} {Mistral 7b}.

\bibitem[{Kao et~al.(2014{\natexlab{a}})Kao, Bergen, and Goodman}]{kao2014formalizing}
Justine Kao, Leon Bergen, and Noah Goodman. 2014{\natexlab{a}}.
\newblock Formalizing the pragmatics of metaphor understanding.
\newblock In \emph{Proceedings of the annual meeting of the Cognitive Science Society}, volume~36.

\bibitem[{Kao and Goodman(2015)}]{Kao2015Irony}
Justine~T. Kao and Noah~D. Goodman. 2015.
\newblock \href {https://api.semanticscholar.org/CorpusID:4693345} {Let's talk (ironically) about the weather: Modeling verbal irony}.
\newblock \emph{Cognitive Science}.

\bibitem[{Kao et~al.(2014{\natexlab{b}})Kao, Wu, Bergen, and Goodman}]{kao_number_words_2014}
Justine~T. Kao, Jean~Y. Wu, Leon Bergen, and Noah~D. Goodman. 2014{\natexlab{b}}.
\newblock \href {https://doi.org/10.1073/pnas.1407479111} {Nonliteral understanding of number words}.
\newblock \emph{Proceedings of the National Academy of Sciences}, 111(33):12002--12007.

\bibitem[{Lai and Nissim(2022)}]{lai-nissim-2022-multi}
Huiyuan Lai and Malvina Nissim. 2022.
\newblock \href {https://aclanthology.org/2022.coling-1.519/} {Multi-figurative language generation}.
\newblock In \emph{Proceedings of the 29th International Conference on Computational Linguistics}, pages 5939--5954, Gyeongju, Republic of Korea. International Committee on Computational Linguistics.

\bibitem[{Lai et~al.(2023)Lai, Toral, and Nissim}]{lai-etal-2023-multilingual}
Huiyuan Lai, Antonio Toral, and Malvina Nissim. 2023.
\newblock \href {https://doi.org/10.18653/v1/2023.findings-acl.589} {Multilingual multi-figurative language detection}.
\newblock In \emph{Findings of the Association for Computational Linguistics: ACL 2023}, pages 9254--9267, Toronto, Canada. Association for Computational Linguistics.

\bibitem[{Li and Sporleder(2010)}]{li-sporleder-2010-using}
Linlin Li and Caroline Sporleder. 2010.
\newblock \href {https://aclanthology.org/N10-1039/} {Using {G}aussian mixture models to detect figurative language in context}.
\newblock In \emph{Human Language Technologies: The 2010 Annual Conference of the North {A}merican Chapter of the Association for Computational Linguistics}, pages 297--300, Los Angeles, California. Association for Computational Linguistics.

\bibitem[{Liu et~al.(2022)Liu, Cui, Zheng, and Neubig}]{liu-etal-2022-testing}
Emmy Liu, Chenxuan Cui, Kenneth Zheng, and Graham Neubig. 2022.
\newblock \href {https://doi.org/10.18653/v1/2022.naacl-main.330} {Testing the ability of language models to interpret figurative language}.
\newblock In \emph{Proceedings of the 2022 Conference of the North American Chapter of the Association for Computational Linguistics: Human Language Technologies}, pages 4437--4452, Seattle, United States. Association for Computational Linguistics.

\bibitem[{Monroe et~al.(2017)Monroe, Hawkins, Goodman, and Potts}]{monroe-etal-2017-colors}
Will Monroe, Robert~X.D. Hawkins, Noah~D. Goodman, and Christopher Potts. 2017.
\newblock \href {https://doi.org/10.1162/tacl_a_00064} {Colors in context: A pragmatic neural model for grounded language understanding}.
\newblock \emph{Transactions of the Association for Computational Linguistics}, 5:325--338.

\bibitem[{Park et~al.(2024)Park, Lee, Park, Jeong, Koo, Hwang, Park, and Lee}]{park-etal-2024-multiprageval}
Dojun Park, Jiwoo Lee, Seohyun Park, Hyeyun Jeong, Youngeun Koo, Soonha Hwang, Seonwoo Park, and Sungeun Lee. 2024.
\newblock \href {https://doi.org/10.18653/v1/2024.genbench-1.7} {{M}ulti{P}rag{E}val: Multilingual pragmatic evaluation of large language models}.
\newblock In \emph{Proceedings of the 2nd GenBench Workshop on Generalisation (Benchmarking) in NLP}, pages 96--119, Miami, Florida, USA. Association for Computational Linguistics.

\bibitem[{Reimers and Gurevych(2019)}]{reimers-gurevych-2019-sentence}
Nils Reimers and Iryna Gurevych. 2019.
\newblock \href {https://doi.org/10.18653/v1/D19-1410} {Sentence-{BERT}: Sentence embeddings using {S}iamese {BERT}-networks}.
\newblock In \emph{Proceedings of the 2019 Conference on Empirical Methods in Natural Language Processing and the 9th International Joint Conference on Natural Language Processing (EMNLP-IJCNLP)}, pages 3982--3992, Hong Kong, China. Association for Computational Linguistics.

\bibitem[{Roberts and Kreuz(1994)}]{robertsWhyPeopleUse1994}
Richard~M. Roberts and Roger~J. Kreuz. 1994.
\newblock \href {https://www.jstor.org/stable/40063089} {Why {Do} {People} {Use} {Figurative} {Language}?}
\newblock \emph{Psychological Science}, 5(3):159--163.
\newblock Publisher: [Association for Psychological Science, Sage Publications, Inc.].

\bibitem[{Scontras et~al.(2025)Scontras, Tessler, and Franke}]{scontras2025rsa}
Gregory Scontras, Michael~Henry Tessler, and Michael Franke. 2025.
\newblock Probabilistic language understanding: An introduction to the rational speech act framework.
\newblock \url{https://www.problang.org}.

\bibitem[{Shen et~al.(2019)Shen, Fried, Andreas, and Klein}]{shenPragmaticallyInformativeText2019}
Sheng Shen, Daniel Fried, Jacob Andreas, and Dan Klein. 2019.
\newblock \href {http://arxiv.org/abs/1904.01301} {Pragmatically {Informative} {Text} {Generation}}.
\newblock ArXiv:1904.01301 [cs].

\bibitem[{Spotorno et~al.(2012)Spotorno, Koun, Prado, Van Der~Henst, and Noveck}]{spotornoNeuralEvidenceThat2012}
Nicola Spotorno, Eric Koun, Jérôme Prado, Jean-Baptiste Van Der~Henst, and Ira~A. Noveck. 2012.
\newblock \href {https://doi.org/10.1016/j.neuroimage.2012.06.046} {Neural evidence that utterance-processing entails mentalizing: {The} case of irony}.
\newblock \emph{NeuroImage}, 63(1):25--39.

\bibitem[{Sravanthi et~al.(2024)Sravanthi, Doshi, Tankala, Murthy, Dabre, and Bhattacharyya}]{sravanthi-etal-2024-pub}
Settaluri Sravanthi, Meet Doshi, Pavan Tankala, Rudra Murthy, Raj Dabre, and Pushpak Bhattacharyya. 2024.
\newblock \href {https://doi.org/10.18653/v1/2024.findings-acl.719} {{PUB}: A pragmatics understanding benchmark for assessing {LLM}s' pragmatics capabilities}.
\newblock In \emph{Findings of the Association for Computational Linguistics: ACL 2024}, pages 12075--12097, Bangkok, Thailand. Association for Computational Linguistics.

\bibitem[{Tsvetkov et~al.(2014)Tsvetkov, Boytsov, Gershman, Nyberg, and Dyer}]{tsvetkov-etal-2014-metaphor}
Yulia Tsvetkov, Leonid Boytsov, Anatole Gershman, Eric Nyberg, and Chris Dyer. 2014.
\newblock \href {https://doi.org/10.3115/v1/P14-1024} {Metaphor detection with cross-lingual model transfer}.
\newblock In \emph{Proceedings of the 52nd Annual Meeting of the Association for Computational Linguistics (Volume 1: Long Papers)}, pages 248--258, Baltimore, Maryland. Association for Computational Linguistics.

\bibitem[{Tsvilodub et~al.(2025)Tsvilodub, Gandhi, Zhao, Fränken, Franke, and Goodman}]{tsvilodub2025nonliteralunderstandingnumberwords}
Polina Tsvilodub, Kanishk Gandhi, Haoran Zhao, Jan-Philipp Fränken, Michael Franke, and Noah~D. Goodman. 2025.
\newblock \href {http://arxiv.org/abs/2502.06204} {Non-literal understanding of number words by language models}.

\bibitem[{Van~Hee et~al.(2018)Van~Hee, Lefever, and Hoste}]{van-hee-etal-2018-semeval}
Cynthia Van~Hee, Els Lefever, and V{\'e}ronique Hoste. 2018.
\newblock \href {https://doi.org/10.18653/v1/S18-1005} {{S}em{E}val-2018 task 3: Irony detection in {E}nglish tweets}.
\newblock In \emph{Proceedings of the 12th International Workshop on Semantic Evaluation}, pages 39--50, New Orleans, Louisiana. Association for Computational Linguistics.

\bibitem[{Wolf et~al.(2020)Wolf, Debut, Sanh, Chaumond, Delangue, Moi, Cistac, Rault, Louf, Funtowicz, Davison, Shleifer, von Platen, Ma, Jernite, Plu, Xu, Scao, Gugger, Drame, Lhoest, and Rush}]{wolf2020huggingfacestransformersstateoftheartnatural}
Thomas Wolf, Lysandre Debut, Victor Sanh, Julien Chaumond, Clement Delangue, Anthony Moi, Pierric Cistac, Tim Rault, Rémi Louf, Morgan Funtowicz, Joe Davison, Sam Shleifer, Patrick von Platen, Clara Ma, Yacine Jernite, Julien Plu, Canwen Xu, Teven~Le Scao, Sylvain Gugger, Mariama Drame, Quentin Lhoest, and Alexander~M. Rush. 2020.
\newblock \href {http://arxiv.org/abs/1910.03771} {Huggingface's transformers: State-of-the-art natural language processing}.

\end{thebibliography}

\clearpage

\appendix

\onecolumn

\section{Additional Theoretical Results and Discussion}

\label{app:rsa_theory}

\subsection{Equivalence of RSA Formulations}

\label{app:equivalence_rsa_formulations}

We demonstrate that the pragmatic speaker formula introduced in Section~\ref{sec:standard_rsa} is equivalent to the formulation originally presented in \citet{frank_ref_game_2012}, which is given by
\begin{align}
    P_{S_1}(u|m,c) \propto \exp(\alpha\cdot U(m,u,c)),
    \label{eq:og_pragmatic_speaker}
\end{align}
where $U(m,u,c)$ is the utility function defined via information theory as
\begin{align}
    U(m,u,c) = \log P_{L_0}(m|c,u) - \kappa(u),
\end{align}
where $\kappa: \mathcal{U} \rightarrow \mathbb{R}_{\geq 0}$ is some cost function. 

Replacing $U(m,u,c)$ in Equation~\ref{eq:og_pragmatic_speaker}, this is equivalent to
\begin{equation}
    P_{S_1}(u|m,c) \propto \exp(-\alpha\cdot\kappa(u)) \cdot P_{L_0}(m|c,u)^\alpha.
    \label{eq:og_simplified}
\end{equation}

Firstly, we show that there exists a particular setting of $\kappa(u)$ in Equation~\ref{eq:og_simplified} for which we can recover Equation~\ref{eq:rsa_pragmatic_speaker} presented in Section~\ref{sec:standard_rsa}:
\begin{align*}
    P_{S_1}(u|m,c) &\propto \exp(-\alpha\cdot\kappa(u)) \cdot P_{L_0}(m|c,u)^\alpha \\
    &= \exp(\log (P(u|c)) \cdot P_{L_0}(m|c,u)^\alpha & \text{Setting }\kappa(u) = \frac{\log P(u|c)}{-\alpha}\\ 
    &= P_{L_0}(m|c,u)^\alpha \cdot P(u|c) 
\end{align*}

Secondly, we show that there exists a particular setting of $P(u|c)$ for which we can recover Equation~\ref{eq:og_simplified}:
\begin{align*}
    P_{S_1}(u|m,c) &\propto P_{L_0}(m|c,u)^\alpha \cdot P(u|c) \\
    &= P_{L_0}(m|c,u)^\alpha \cdot \texttt{softmax}(-\alpha\cdot\kappa(u)) & \text{Setting }P(u|c) = \texttt{softmax}(-\alpha\cdot\kappa(u)) \\ 
    &\propto P_{L_0}(m|c,u)^\alpha \cdot \exp(-\alpha\cdot\kappa(u))
\end{align*}

This proves the equivalence.

\subsection{Standard RSA Provides Zero Probability Mass for Non-Literal Interpretations}

\label{app:standard_rsa_non_literal}

We show that in the standard RSA framework, if $\alpha > 0$, then for all $m \in \mathcal{M}$ and $u \in \mathcal{U}$, the condition $m \notin \llbracket u \rrbracket$ implies that $P_{L_1}(m|u) = 0$.  

Let $\alpha > 0$ and consider $m \in \mathcal{M}$ and $u \in \mathcal{U}$ such that $m \notin \llbracket u \rrbracket$. Then, we have:  
\begin{align*}
    P_{L_0}(m|c,u) &\propto \mathds{1}_{m \in \llbracket u \rrbracket} \cdot P(m|c) \\ 
    &= 0 \cdot P(m|c) \\ 
    &= 0, \\
    P_{S_1}(u|c,m) &\propto P_{L_0}(m|c,u)^\alpha \cdot P(u|c) \\ 
    &= 0 \cdot P(u|c) \\ 
    &= 0, \\
    P_{L_1}(m|c,u) &\propto P_{S_1}(u|c,m) P(m|c) \\ 
    &= 0 \cdot P(m|c) \\ 
    &= 0.
\end{align*}
This proves the above statement.

\subsection{QUD-RSA is a Special Case of \rsatwo}

\label{app:qud_rsa_special_case}

We show that affect-aware RSA is a special case of \rsatwo. To do so, we use the more general formulation of affect-aware RSA, Question Under Discussion RSA (QUD-RSA) \citep{scontras2025rsa}.

In QUD-RSA, the meaning space $\mathcal{M}$ is projected to meaning subspaces $\mathcal{X}$ using projection functions $q : \mathcal{M} \rightarrow \mathcal{X}$. Typically, this is done when the meaning space has been augmented to a vector space in order to include additional meaning dimensions. For instance, in affect-aware RSA, the meaning vector space $\mathcal{M} = \mathcal{S} \times \mathcal{A}$ is broken into the state of the world being conveyed, $\mathcal{S}$, and the affect $\mathcal{A}$. In this setting, an affect projection $q_\texttt{affect} : \mathcal{S} \times \mathcal{A} \rightarrow \mathcal{A}$ might look like the following:
\begin{equation*}
    q_\texttt{affect}(s,a) = a.
\end{equation*}

We first define the equations of QUD-RSA and of \rsatwospace and then show that any instance of the QUD-RSA equations can be re-written with the \rsatwospace equations, but not vice-versa.

\subsubsection{Definitions}

\textbf{Definition 1 (QUD-RSA).} Given a meaning space $\mathcal{M}$, an utterance space $\mathcal{U}$, a context space $\mathcal{C}$ and a set of $n$ projection functions $Q = \{q_i : i \in [1\dots n] \And q_i \text{ is a projection function from }\mathcal{M}\text{ to some subspace }\mathcal{X}\}$, the QUD-RSA equations $\mathcal{E}_\texttt{QUD-RSA}$, are as follows:\footnote{We overload notation here and use $q$ to represent both the latent variable which indexes its corresponding projection \emph{as well as} the projection function itself.}
\begin{align}
P_{L_0}(m \mid u, c) &\propto \mathds{1}_{m \in \llbracket u \rrbracket} \cdot P(m \mid c), \\
P_{L_0}(m \mid c, u, q) &\propto \sum_{m' \in \mathcal{M}} \mathds{1}_{q(m') = q(m)} \cdot P_{L_0}(m' \mid u, c), \label{eq:qud_rsa_literal}\\
P_{S_1}(u \mid c, m, q) &\propto P_{L_0}(m \mid c, u, q) \cdot P(u \mid c), \label{quds1}\\
P_{L_1}(m \mid c, u, q) &\propto P_{S_1}(u \mid c, m, q) \cdot P(m \mid c). \label{qudl1}
\end{align}

\vspace{0.5cm}

\noindent\textbf{Definition 2 (\rsatwo).} Given a meaning space $\mathcal{M}$, an utterance space $\mathcal{U}$, a context space $\mathcal{C}$, a rhetorical strategy space $\mathcal{R}$ and a set of rhetorical strategy functions $F_r = \{f_r: r \in \mathcal{R} \And f_r: \mathcal{M} \times \mathcal{C} \times \mathcal{U} \rightarrow [0,1]\}$, the \rsatwospace equations, $\mathcal{E}_{\rsatwo}$ are as follows:
\begin{align}
P_{L_0}(m \mid c, u, r) &\propto f_r(c, m, u) \cdot P(m \mid c), \label{eq:gen_rsa_pl0}\\
P_{S_1}(u \mid c, m, r) &\propto P_{L_0}(m \mid c, u, r) \cdot P(u \mid c), \\
P_{L_1}(m \mid c, u, r) &\propto P_{S_1}(u \mid c, m, r) \cdot P(m \mid c).
\end{align}

\subsubsection{QUD-RSA Can Be Simulated by \rsatwo}

\noindent\textbf{Lemma 1.} If $P(m \mid c) > 0$ for all $m \in \mathcal{M}, c \in \mathcal{C}$, any instance of the QUD-RSA equations can be represented as an instance of the \rsatwospace equations.

\textbf{Proof.} Given an instance of the QUD-RSA equations, $\mathcal{E}_\texttt{QUD-RSA}$, we can build an equivalent instance of the \rsatwospace equations, $\mathcal{E}_{\rsatwo}$. 

We do this by setting the spaces $\mathcal{C}, \mathcal{M} \And \mathcal{U}$ in $\mathcal{E}_{\rsatwo}$ to be the same as the ones from $\mathcal{E}_\texttt{QUD-RSA}$. In addition, we define a rhetorical strategy variable $r$ and a corresponding rhetorical function $f_r$ for each projection variable and corresponding projection function $q$ of $\mathcal{E}_\texttt{QUD-RSA}$ as follows:
\begin{equation}
f_r(m, c, u) = \frac{ \sum_{m' \in \mathcal{M}} \mathds{1}_{q(m') = q(m)} \cdot P_{L_0}(m' \mid c, u) }{ k \cdot P(m \mid c) },
\end{equation}
where the division by $k = \max_{m'}f_r(c,m',u)$ is needed such that $f_r(c,m,u) \in [0,1]$.

Replacing $f_r$ in $\mathcal{E}_{\rsatwo}$'s Equation~\ref{eq:gen_rsa_pl0} enables us to recover the instance of the QUD-RSA equations, $\mathcal{E}_\texttt{QUD-RSA}$.

\vspace{0.5cm}

\subsubsection{Not All Instances of \rsatwospace Can Be Simulated by QUD-RSA}

\noindent\textbf{Lemma 2.} Given context space $\mathcal{C}$, meaning space $\mathcal{M}$, and utterance space $\mathcal{U}$, there exists an instance of the \rsatwospace equations which cannot be simulated by any instance of the QUD-RSA equations.

\textbf{Proof Idea.} The crux of this proof lies in the observation that the QUD-RSA equations can only induce literal listener distributions which are \emph{binary combinations} of meaning priors $P(m|c)$. Thus, one can pick some probability vector which is not such a combination and compute it using the rhetorical strategy function $f_r$.

\textbf{Proof.} We first demonstrate that the literal listener of any instance of the QUD-RSA equations is a binary combination of meaning priors.

Consider the normalized literal listener distribution $P_{L_0}(m \mid c, u)$ from the QUD-RSA equations:
\begin{equation}
P_{L_0}(m \mid c, u) = 
\frac{\mathds{1}_{m \in \llbracket u \rrbracket} \cdot P(m \mid c) }
     { \sum\limits_{m''} \mathds{1}_{m'' \in \llbracket u \rrbracket} \cdot P(m'' \mid c) }.
\end{equation}
Replacing it in Equation~\ref{eq:qud_rsa_literal}, we get the following:
\begin{align}
    P_{L_0}(m \mid c, u, q) &\propto \sum_{m'} \mathds{1}_{q(m') = q(m)} \cdot \Bigl( \frac{\mathds{1}_{m \in \llbracket u \rrbracket} \cdot P(m \mid c) }{ \sum\limits_{m''} \mathds{1}_{m'' \in \llbracket u \rrbracket} \cdot P(m'' \mid c) } \Bigr) \\
    &\propto \sum_{m'} \mathds{1}_{q(m') = q(m)} \cdot \mathds{1}_{m \in \llbracket u \rrbracket} \cdot P(m \mid c) \\
    &\propto \sum_{m'} \mathds{1}_{q(m') = q(m) \wedge m \in \llbracket u \rrbracket} \cdot P(m \mid c)
\end{align}

Thus, the QUD-RSA literal listeners are probability distributions induced by taking binary combinations of the meaning priors, $P(m|c)$. Since the meaning space $\mathcal{M}$ is finite, this implies that there is a finite set of distributions which the QUD-RSA literal listener can be set to. We define this set as follows:
\begin{align}
    \mathcal{P_{L}}_0 = \{ p_{l_0} \in \mathbb{R}^{|\mathcal{M}|} : p_{l_0}\text{ is a QUD-RSA literal listener probability vector}\}.
\end{align}
Then, to complete this proof, we need only define a function $f_r$ such that $P_{L_0}(m \mid c, u, r) \notin \mathcal{P_{L}}_0$.

We create an $f_r$ which induces $P_{L_0}(m \mid c, u, r)$ outside of $\mathcal{P_{L}}_0$ by picking a real number $k$ such that it lies between $0$ and the minimum non-zero probability value in $\mathcal{P_{L}}_0$. That is,
\begin{align}
k = \frac{p_\text{min}}{2} \text{ such that } p_\text{min} = \min_{\vec{p} \in \mathcal{P_{L}}_0, \, i \in 1 \dots |\mathcal{M}|, \,\vec{p}[i] > 0} \vec{p}[i]
\end{align}

We define $f_r$ such that $P_{L_0}(m \mid c, u, r) = k$ for some $m \in \mathcal{M}$. Let $m_1, m_2 \in \mathcal{M}, m_1 \neq m_2$ and $f_r$ be defined as follows:
\[
f_r(m, c, u) = 
\begin{cases}
\frac{k}{P(m_1 \mid c)} & \text{if } m = m_1 \\
\frac{1 - k}{P(m_2 \mid c)} & \text{if } m = m_2 \\
0 & \text{otherwise}
\end{cases}
\]

With this function $f_r$\footnote{$f_r$ should be divided by its maximum value to ensure it respects its $[0,1]$ co-domain constraint. We omit this step for compactness.}, we can show that $P_{L_0}(m_1 \mid c, u, r) = k$ which we have just shown is not a probability value in any of the probability vectors found in $\mathcal{P_{L}}_0$. This can be seen as follows:
\begin{align}
P_{L_0}(m_1 \mid c, u, r) 
&=  \frac{f_r(m_1, c, u) \cdot P(m_1 \mid c)}{\sum_{m'} f_r(m', c, u) \cdot P(m' \mid c)}, \\
&= \frac{ \frac{k}{P(m_1 \mid c)} \cdot P(m_1 \mid c) }
         { \frac{k}{P(m_1 \mid c)} \cdot P(m_1 \mid c) + \frac{1 - k}{P(m_2 \mid c)} \cdot P(m_2 \mid c) }, \\
&= \frac{k}{k + 1 - k} = k \quad \text{as desired}.
\end{align}

This shows that there exists an instance of the \rsatwospace equations which cannot be represented with QUD-RSA.

\section{Deriving Non-Literal Interpretations of Figurative Language with \rsatwo}

\label{app:affect_aware_rsa_exp}

We provide additional details regarding the ironic weather utterances experiment along with additional figures for both our non-literal number expression and ironic weather utterances experiments.

\subsection{Ironic Weather Utterances Experimental Details}

\subsubsection{Weather Contexts Images}

\label{app:weather_images}

We include the images associated with each of the 9 weather contexts in the ironic weather utterances dataset. These images were shown to human participants to elicit both prior and posterior probabilities. These can also be found in the original work by \citet{Kao2015Irony}.

\begin{figure}[H]
    \centering
    \includegraphics[width=\linewidth]{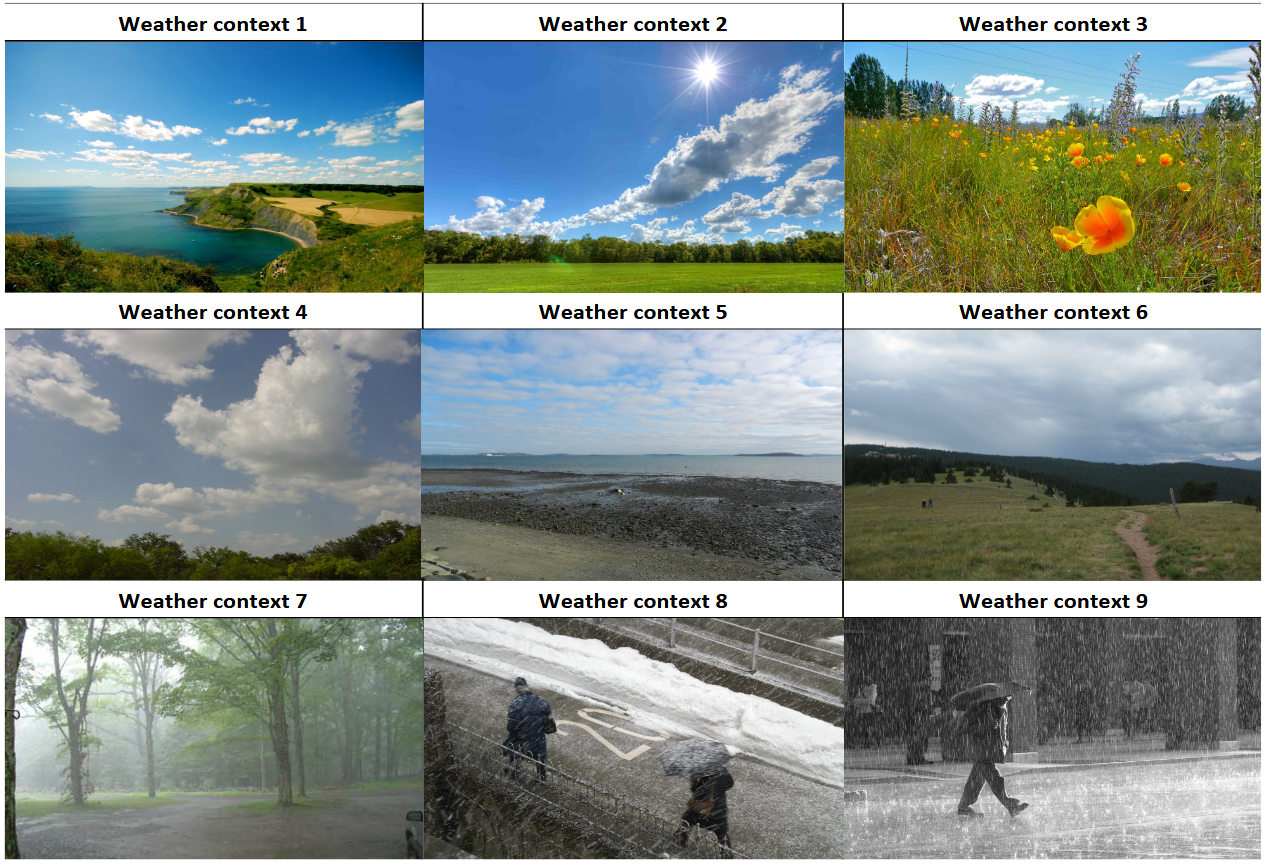}
    \caption{Images associated with the 9 weather contexts in the ironic weather utterances dataset.}
    \label{fig:weather_images}
\end{figure}

\subsubsection{Re-Implementation Details of Affect-Aware RSA}

\label{app:reimplementation_details}

Using the ironic weather utterances dataset from \citet{Kao2015Irony}, we re-implement their affect-aware RSA model using the following equations where $s \in \mathcal{S}$ represents the conveyed state of the world (which we typically call $m$ in our notation), $a, v \in \mathcal{A}, \mathcal{V}$ represent the arousal and valence dimensions of affect respectively and where the three QUD projections used are $q_\texttt{literal}(s,a,v) = s, q_\texttt{arousal}(s,a,v) = a, q_\texttt{valence}(s,a,v) = v$:

\begin{align}
P(s,a,v|c) &= P(s|c) \cdot P(a|c) \cdot P(v|c) \\
P_{L_0}(s,a,v|c,u) &\propto P(s,a,v|c) \cdot \mathds{1}_{s \in \llbracket u \rrbracket}, \\
P_{L_0}(s,a,v|c,u,q) &\propto \sum_{(s',a',v') \in \mathcal{S} \times \mathcal{A} \times \mathcal{V}} \mathds{1}_{q(s',a',v') = q(s,a,v)} \cdot P_{L_0}(s',a',v'|c,u), \\
P_{S_1}(u|c,s,a,v,q) &\propto P_{L_0}(s,a,v|c,u,q)^\alpha \\
L_1(s,a,v|c,u,q) &= P(s,a,v|c) \cdot P_{S_1}(u|c,s,a,v,q) \label{eq:L1_def}
\end{align}

The QUD variables can be marginalized out from either listener with the following equations:

\begin{align}
P_{L_0}(s,a,v|c,u) &= \sum_{q \in Q} P_{L_0}(s,a,v|c,u,q) \cdot P(q|c) \label{eq:PL0_marginalized}, \\
P_{L_1}(s,a,v|c,u) &\propto \sum_{q \in Q} L_1(s,a,v|c,u,q) \cdot P(q|c) \label{eq:PL1_marginalized},
\end{align}

where, as described in \citet{Kao2015Irony}, we use $P(q_\texttt{literal}) = 0.3, P(q_\texttt{arousal}) = 0.4, P(q_\texttt{valence}) = 0.3$.

Finally, we can marginalize out the affect variables $a$ and $v$ from the $P_{L_i}(s,a,v|c,u)$ to get $P_{L_i}(s|c,u)$:

\begin{align}
    P_{L_i}(s|c,u) = \sum_{a \in \mathcal{A}} \sum_{v \in \mathcal{V}} P_{L_i}(s,a,v|c,u) \label{eq:PLi_affect_marginalized}
\end{align}

\begin{figure}
    \centering
    \includegraphics[width=\linewidth]{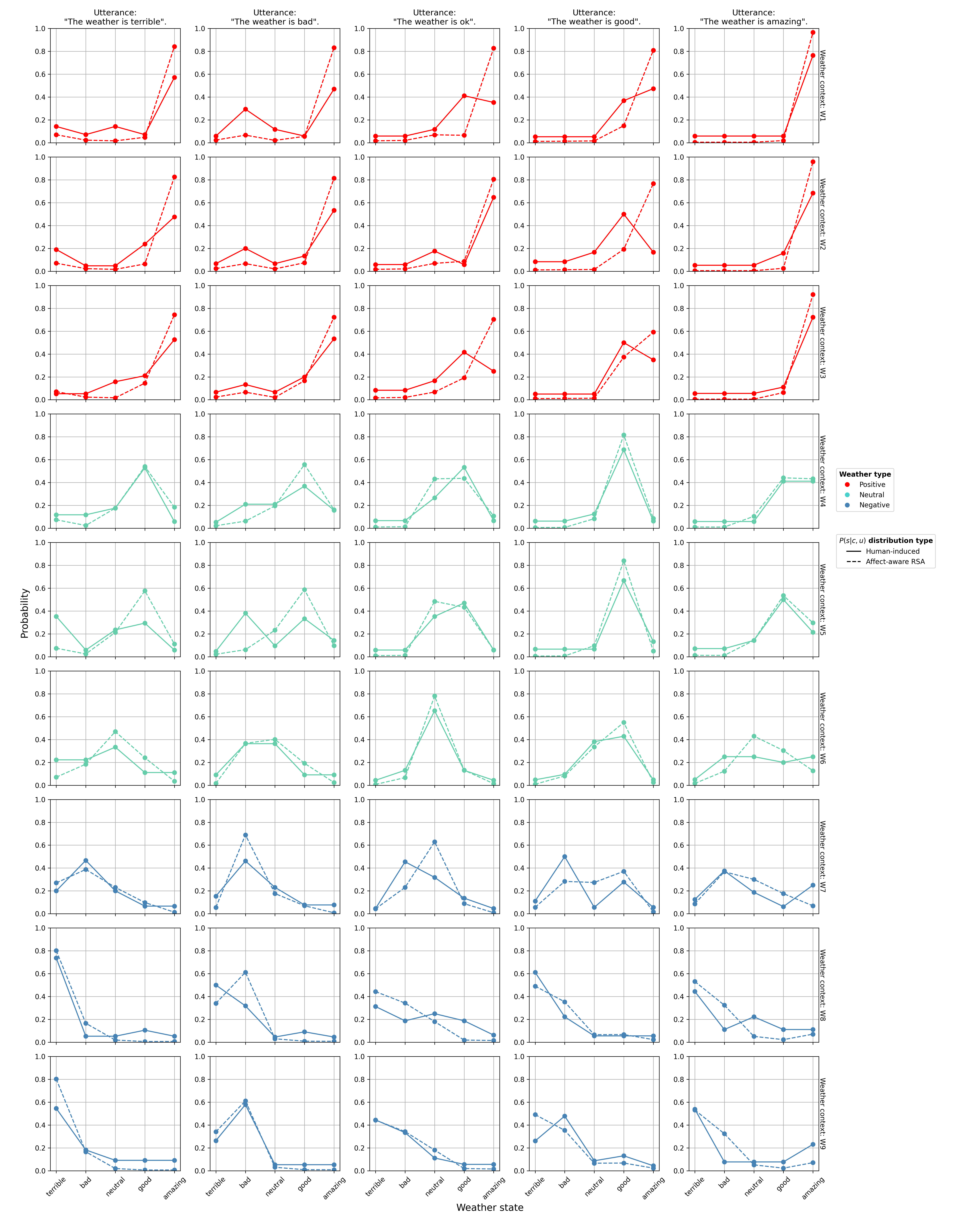}
    \caption{Meaning distributions, $P(m|c,u)$, for humans and for the affect-aware RSA method. This figure was originally generated by \citet{Kao2015Irony}. We reproduce it here with our re-implementation to verify the correctness of the re-implementation.}
    \label{fig:reproduction_kao}
\end{figure}

To verify the correctness of our re-implementation, we reproduce Fig.~5 from the \citet{Kao2015Irony} paper and verify qualitatively that each context-utterance-meaning triple matches. Our reproduction of this figure can be found in Fig.~\ref{fig:reproduction_kao}.

\subsubsection{\rsatwospace Training Details}

To learn the rhetorical strategy function $f_r$, we trained a neural network with an architecture of $16 \times 16 \times 5$, employing sigmoid activation functions throughout. The input to the network was a $16 \times 1$ one-hot encoding vector, where the first 9 entries were reserved for the context indicator ($c_1$ to $c_9$), the next 5 entries were reserved for the utterance indicator, and the final 2 entries were reserved for the rhetorical strategy indicator. The network's output was a $5 \times 1$ vector, representing the values of $f_r$ for each of the five meanings. The use of sigmoid activations ensures that the output values are constrained within the interval $[0,1]$, thereby respecting the defined image of $f_r$. For the training process, the entire dataset was utilized as a single batch, and training proceeded for 500 epochs. We employed a strategy of saving the model that achieved the best performance, as evaluated by the validation loss. The Adam optimizer was chosen for optimization, configured with a learning rate of $0.001$ and a weight decay of $0.001$. The network was trained using a cross-entropy loss function with $P_{L_1}(m|c,u)$.

\label{app:ironic_weather_utterances_exp_details}

\subsection{Non-Literal Number Expressions}

\label{app:non_literal_number_expressions}

\begin{figure}
    \centering
    \includegraphics[width=0.6\linewidth]{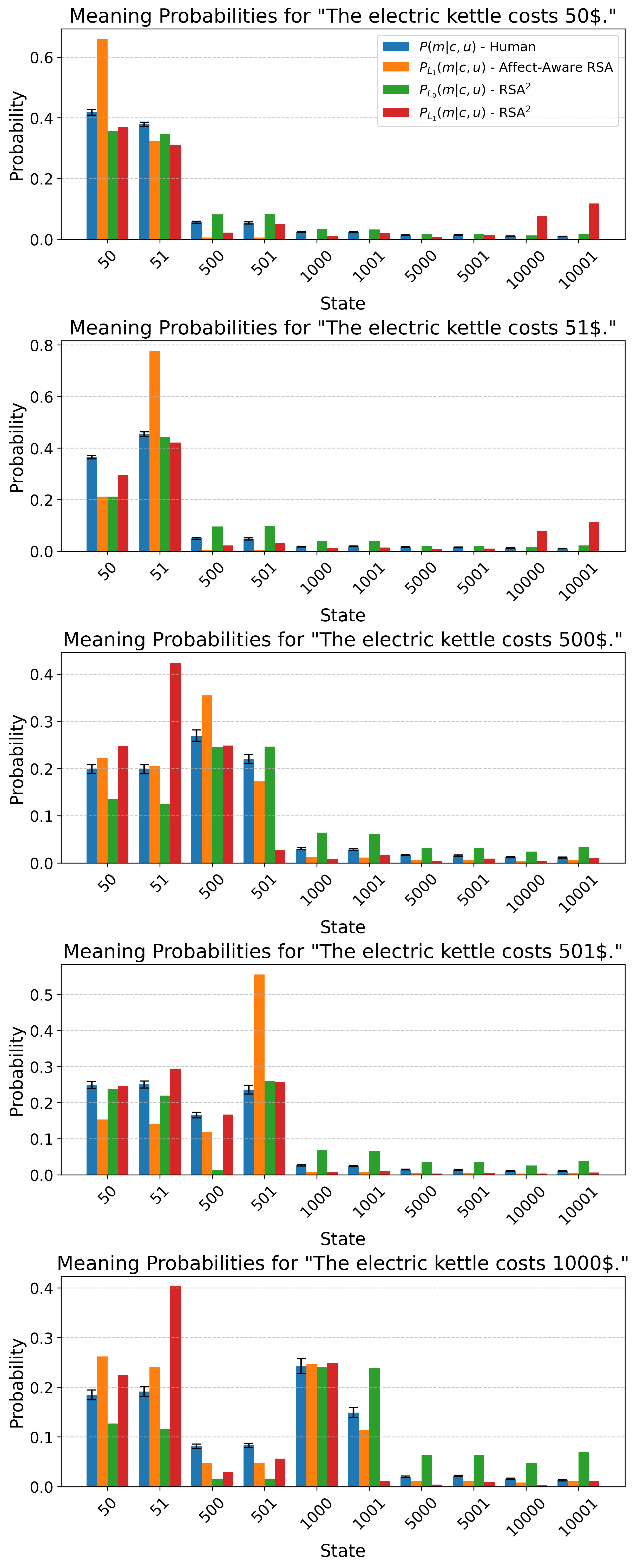}
    \caption{Meaning probability distributions by humans along with the affect-aware pragmatic listener and the \rsatwospace literal and pragmatic listeners on utterances about the price of an electric kettle.}
    \label{fig:electric_kettle_split_0}
\end{figure}

\begin{figure}
    \centering
    \includegraphics[width=0.6\linewidth]{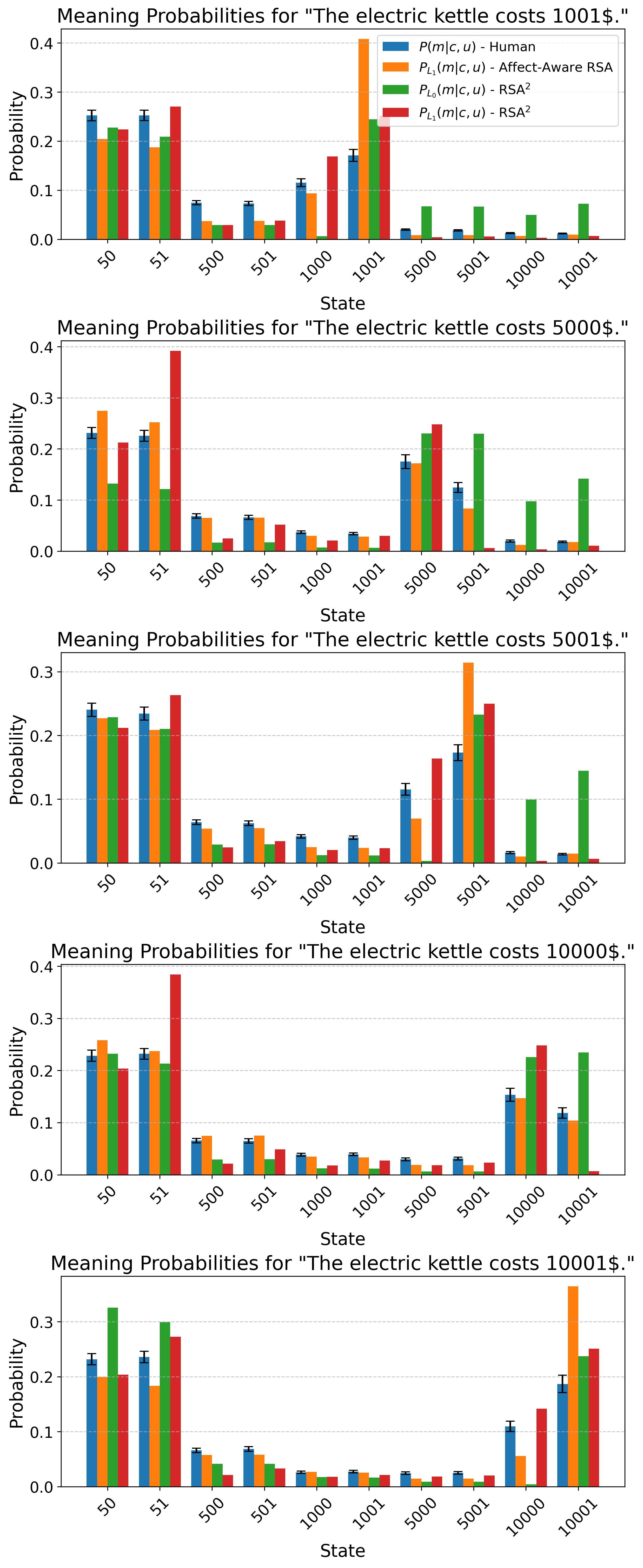}
    \caption{Meaning probability distributions by humans along with the affect-aware pragmatic listener and the \rsatwospace literal and pragmatic listeners on utterances about the price of an electric kettle (Continued).}
    \label{fig:electric_kettle_split_1}
\end{figure}

\begin{figure}
    \centering
    \includegraphics[width=0.6\linewidth]{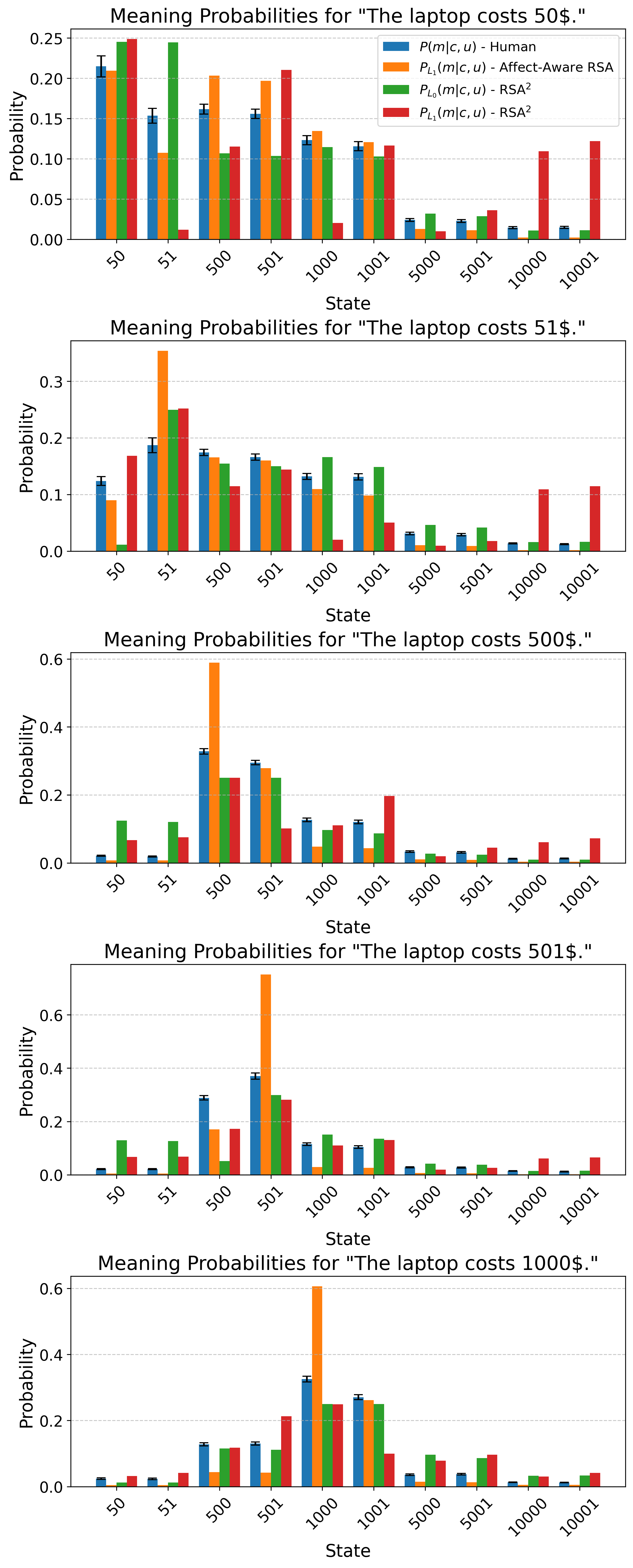}
    \caption{Meaning probability distributions by humans along with the affect-aware pragmatic listener and the \rsatwospace literal and pragmatic listeners on utterances about the price of a laptop.}
    \label{fig:laptop_split_0}
\end{figure}

\begin{figure}
    \centering
    \includegraphics[width=0.6\linewidth]{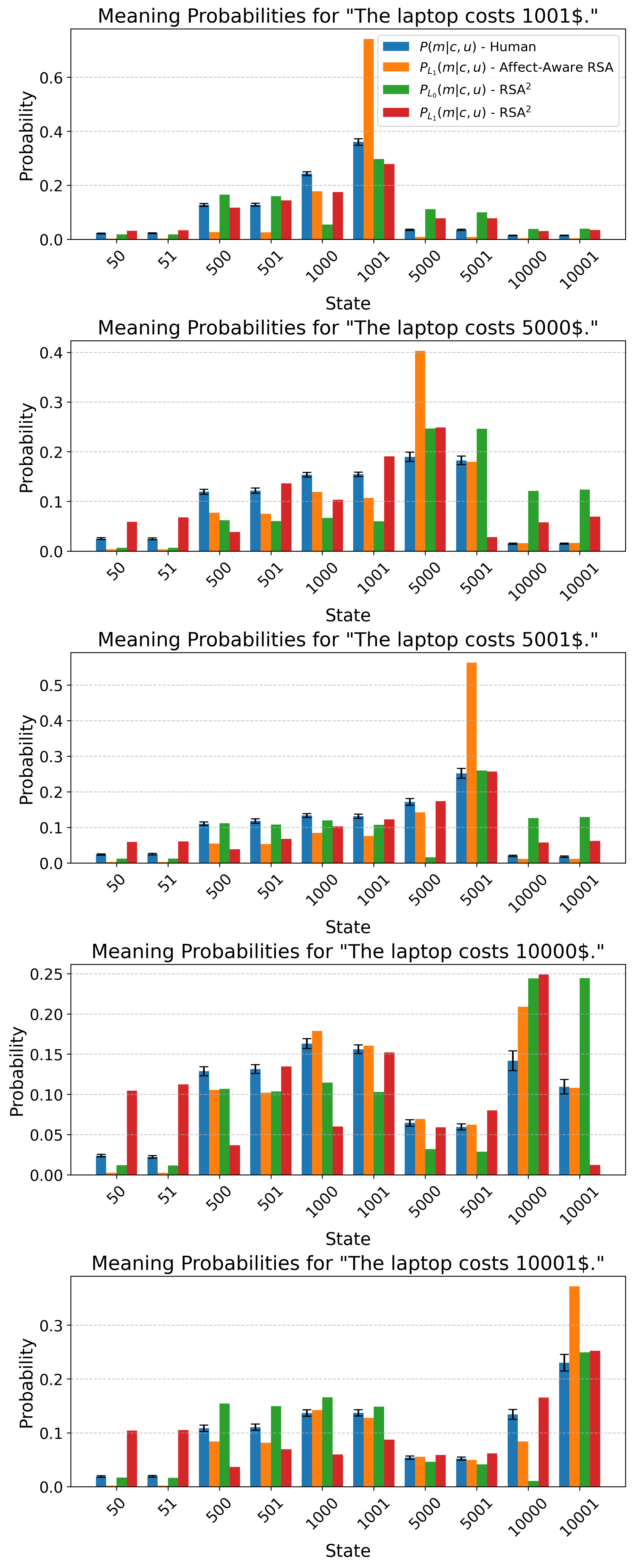}
    \caption{Meaning probability distributions by humans along with the affect-aware pragmatic listener and the \rsatwospace literal and pragmatic listeners on utterances about the price of a laptop (Continued).}
    \label{fig:laptop_split_1}
\end{figure}

\begin{figure}
    \centering
    \includegraphics[width=0.6\linewidth]{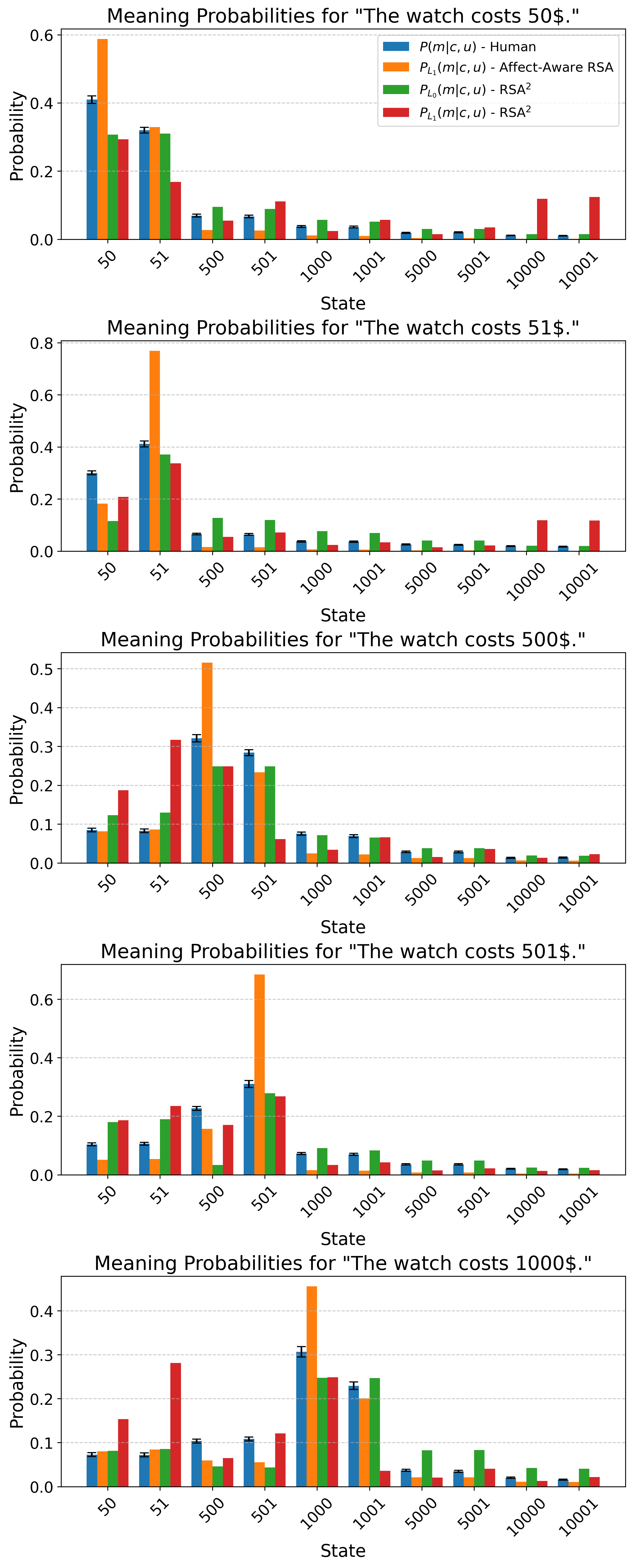}
    \caption{Meaning probability distributions by humans along with the affect-aware pragmatic listener and the \rsatwospace literal and pragmatic listeners on utterances about the price of a watch.}
    \label{fig:watch_split_0}
\end{figure}

\begin{figure}
    \centering
    \includegraphics[width=0.6\linewidth]{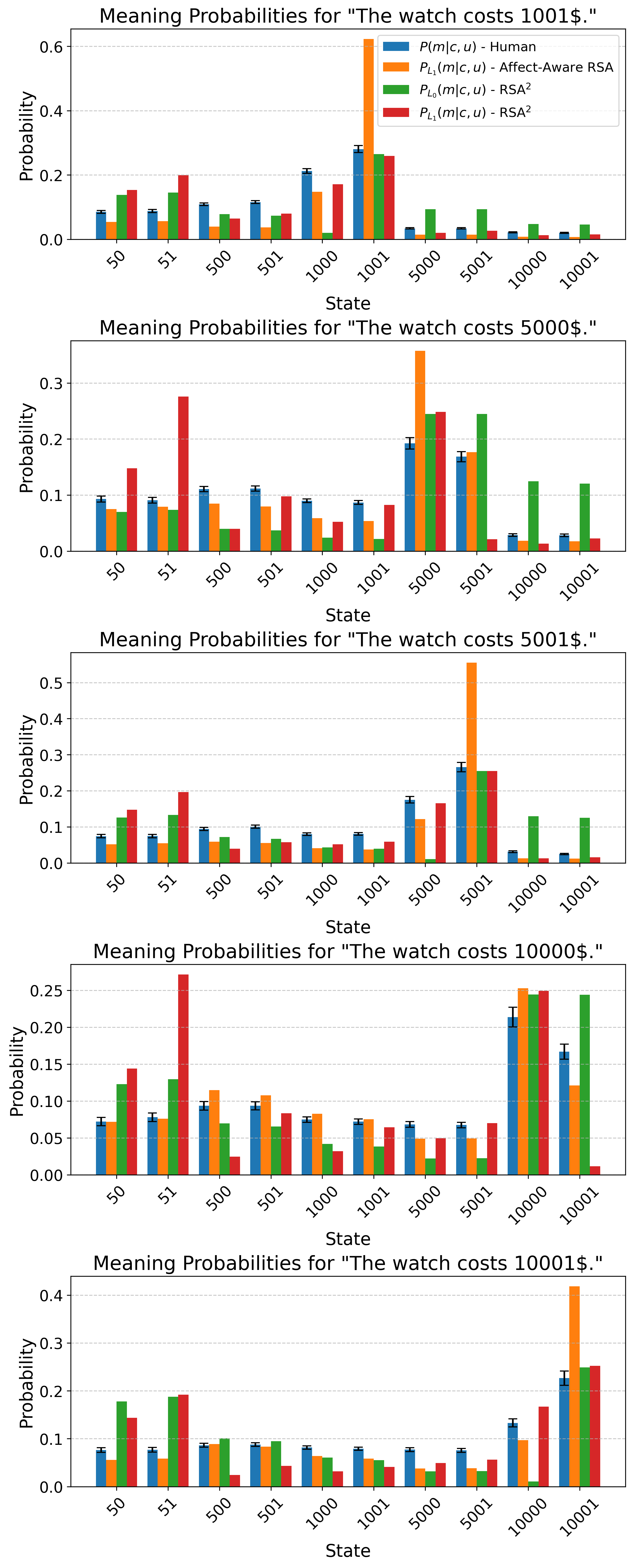}
    \caption{Meaning probability distributions by humans along with the affect-aware pragmatic listener and the \rsatwospace literal and pragmatic listeners on utterances about the price of a watch (Continued).}
    \label{fig:watch_split_1}
\end{figure}

We show the listener meaning distributions for all context-meaning-utterances triples from the non-literal number expressions experiment. Figures \ref{fig:electric_kettle_split_0} and \ref{fig:electric_kettle_split_1} present these distributions for the electric kettle, Figures \ref{fig:laptop_split_0} and \ref{fig:laptop_split_1} present the distributions for the laptop and Figures \ref{fig:watch_split_0} and \ref{fig:watch_split_1} do the same for the watch.

\subsection{Ironic Weather Utterances Additional Results}

\label{app:ironic_weather_utterances_add_results}

\begin{figure}
    \centering
    \includegraphics[width=0.6\linewidth]{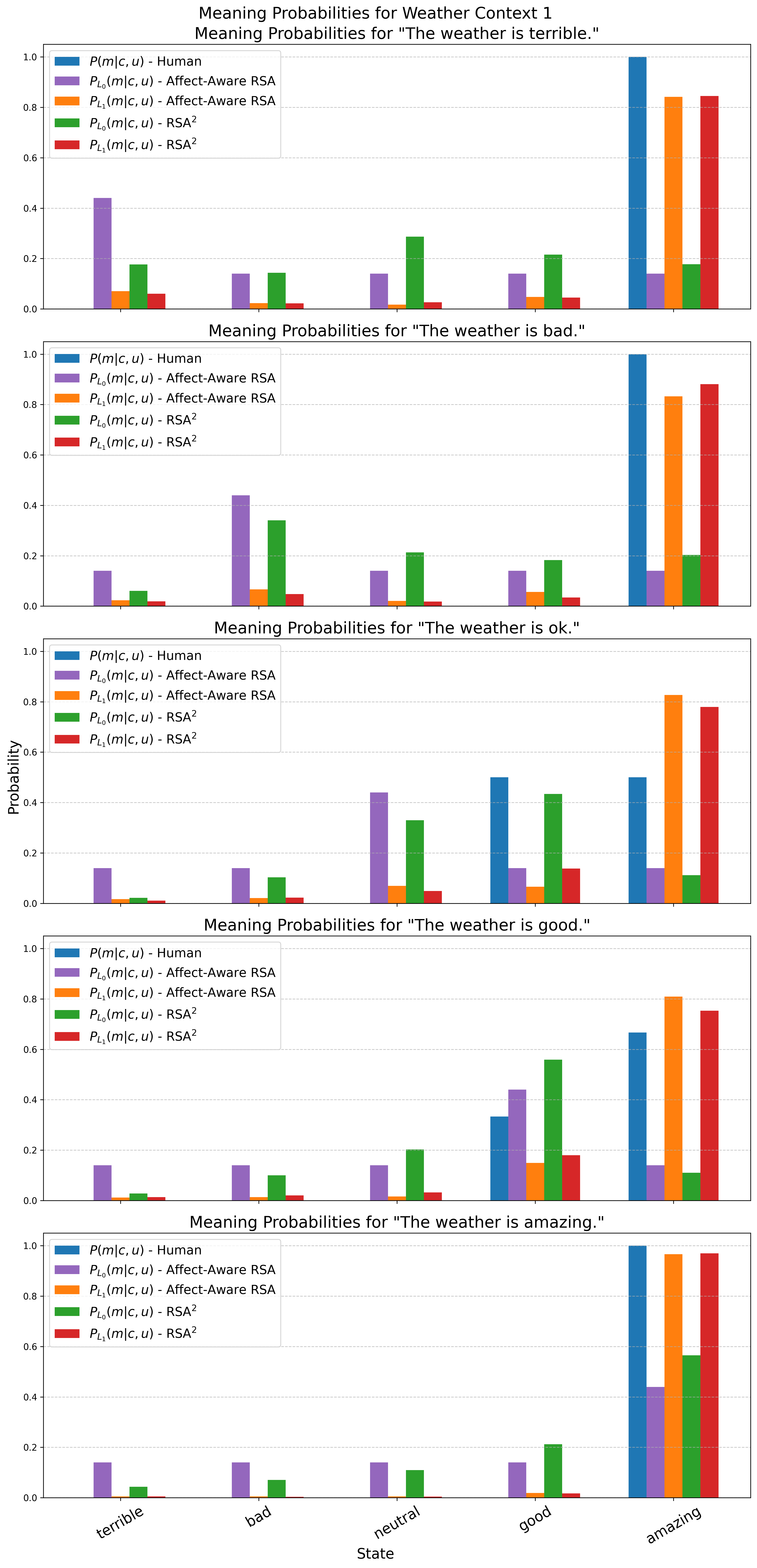}
    \caption{Meaning probability distributions by humans along with the listener and pragmatic listeners of both affect-aware RSA and \rsatwospace for weather context 1 from Fig.~\ref{fig:weather_images}.}
    \label{fig:Weather_context_1}
\end{figure}

\begin{figure}
    \centering
    \includegraphics[width=0.6\linewidth]{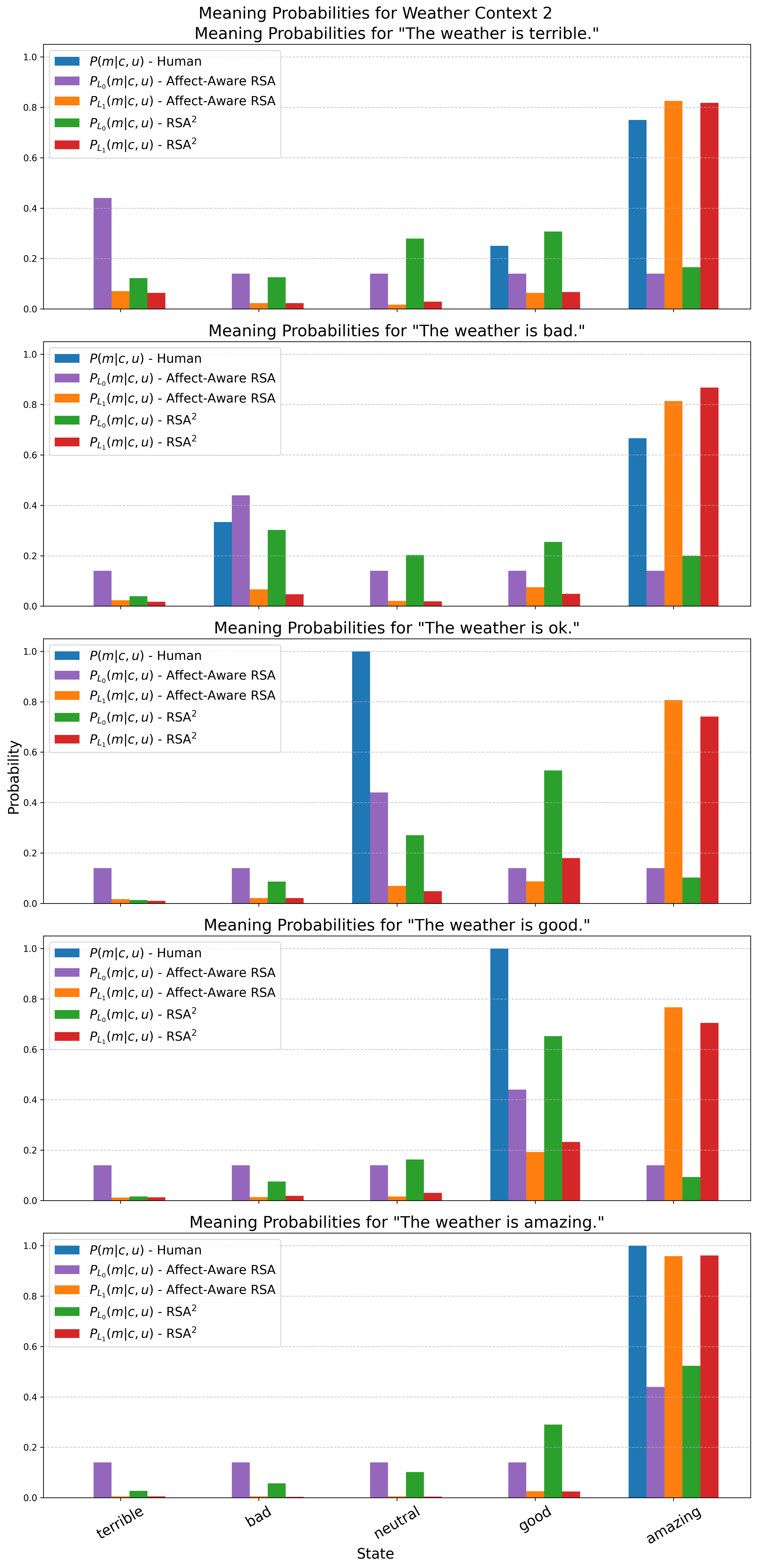}
    \caption{Meaning probability distributions by humans along with the listener and pragmatic listeners of both affect-aware RSA and \rsatwospace for weather context 2 from Fig.~\ref{fig:weather_images}.}
    \label{fig:Weather_context_2}
\end{figure}

\begin{figure}
    \centering
    \includegraphics[width=0.6\linewidth]{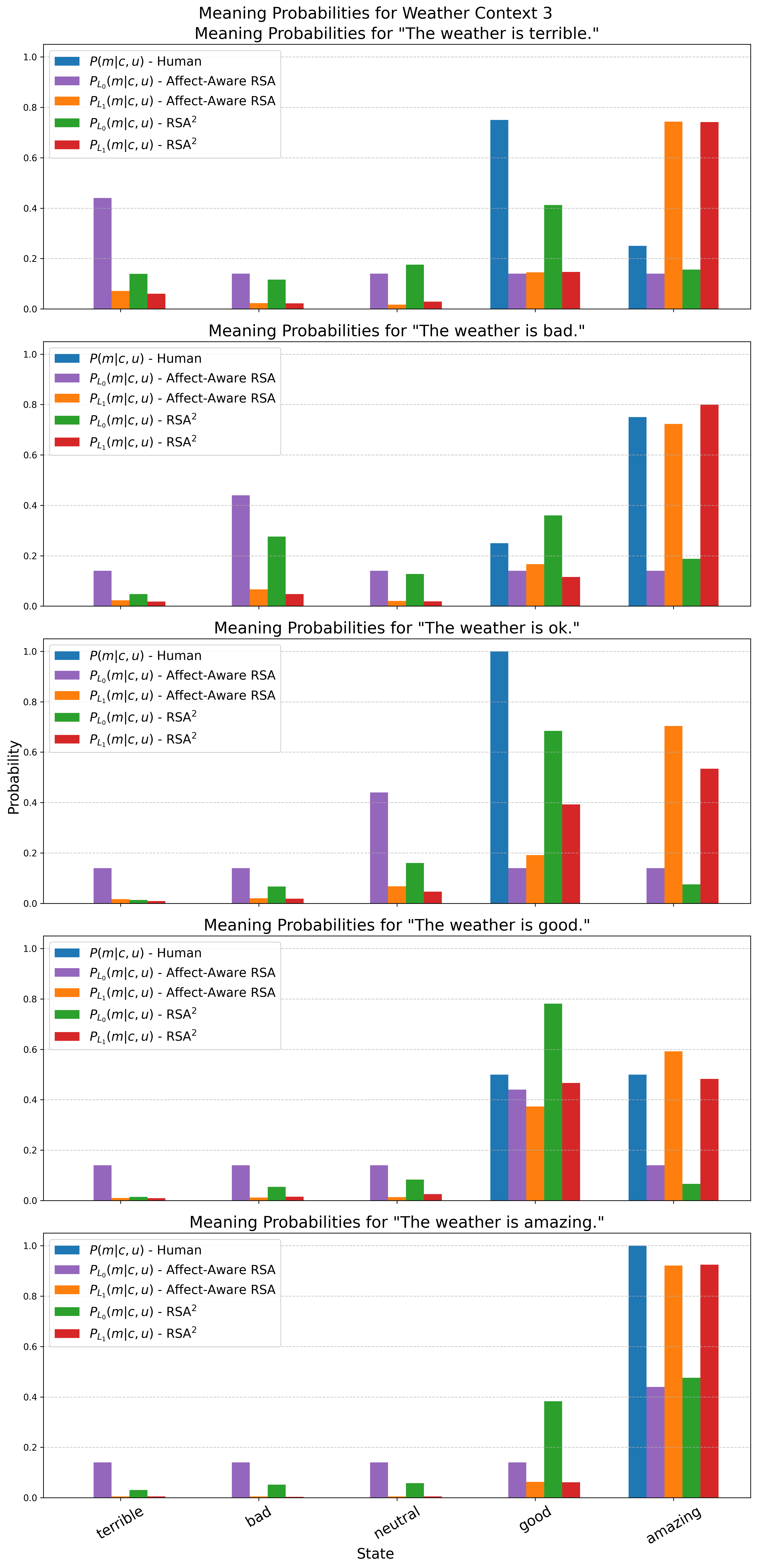}
    \caption{Meaning probability distributions by humans along with the listener and pragmatic listeners of both affect-aware RSA and \rsatwospace for weather context 3 from Fig.~\ref{fig:weather_images}.}
    \label{fig:Weather_context_3}
\end{figure}

\begin{figure}
    \centering
    \includegraphics[width=0.6\linewidth]{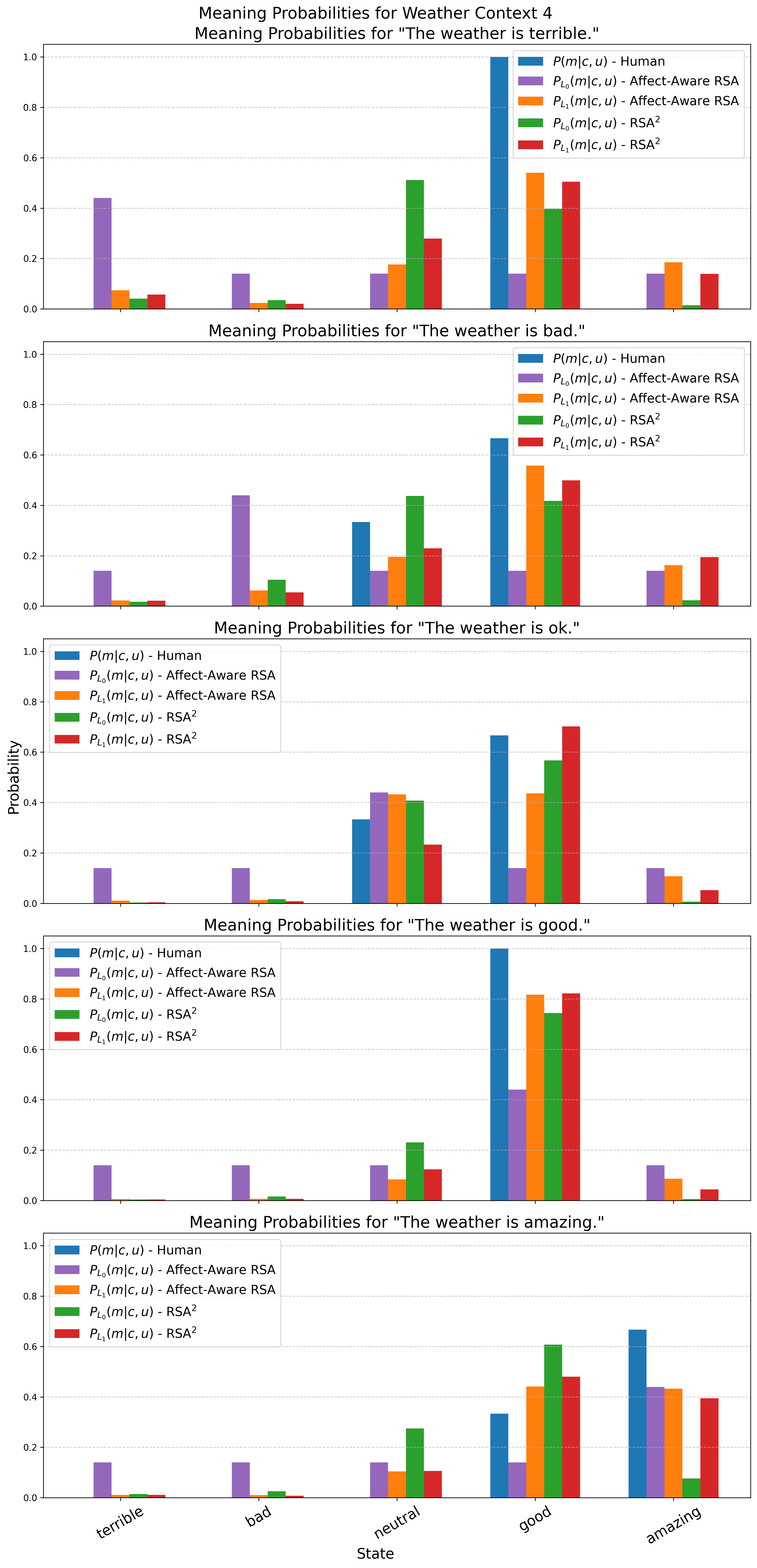}
    \caption{Meaning probability distributions by humans along with the listener and pragmatic listeners of both affect-aware RSA and \rsatwospace for weather context 4 from Fig.~\ref{fig:weather_images}.}
    \label{fig:Weather_context_4}
\end{figure}

\begin{figure}
    \centering
    \includegraphics[width=0.6\linewidth]{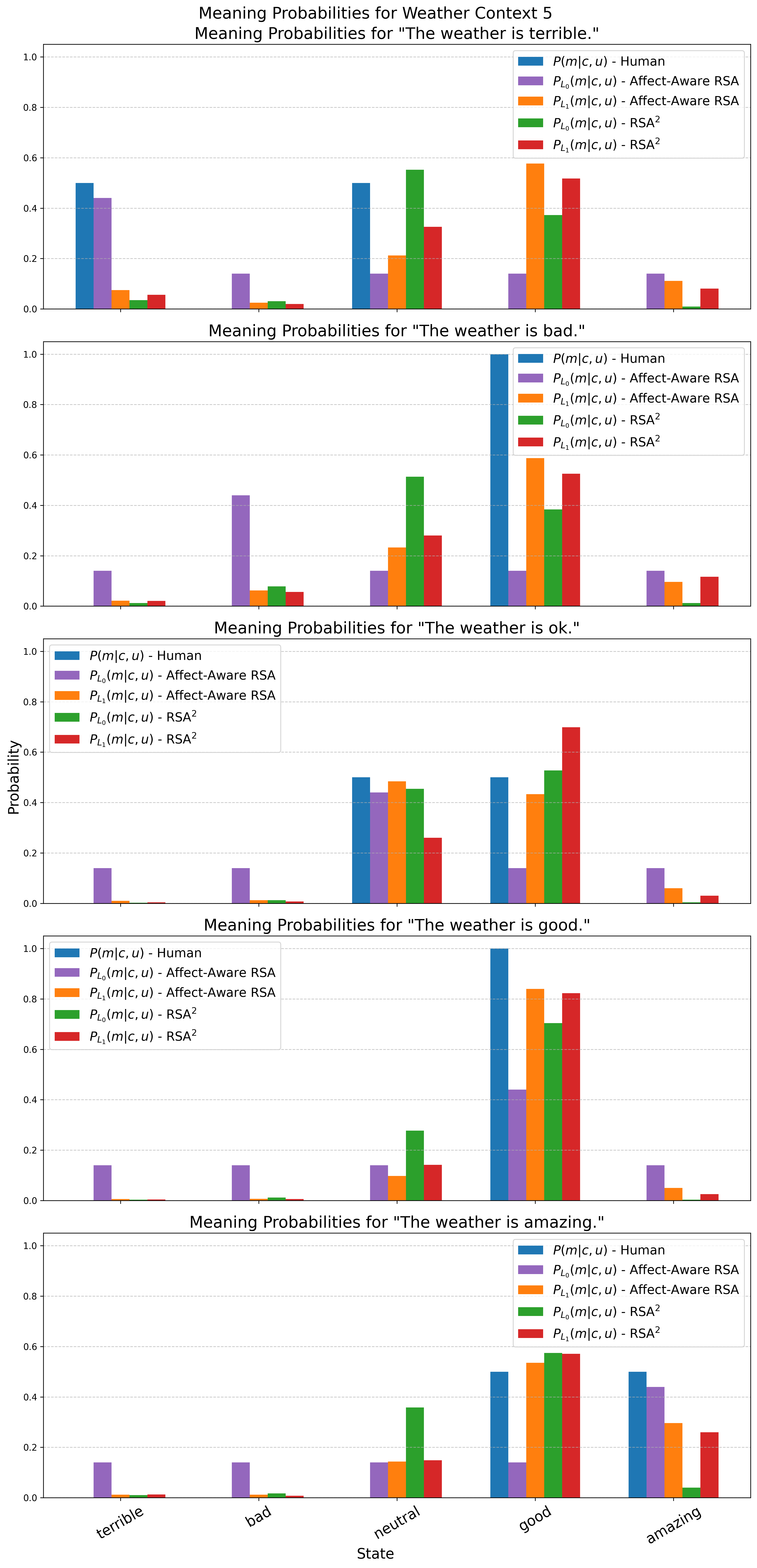}
    \caption{Meaning probability distributions by humans along with the listener and pragmatic listeners of both affect-aware RSA and \rsatwospace for weather context 5 from Fig.~\ref{fig:weather_images}.}
    \label{fig:Weather_context_5}
\end{figure}

\begin{figure}
    \centering
    \includegraphics[width=0.6\linewidth]{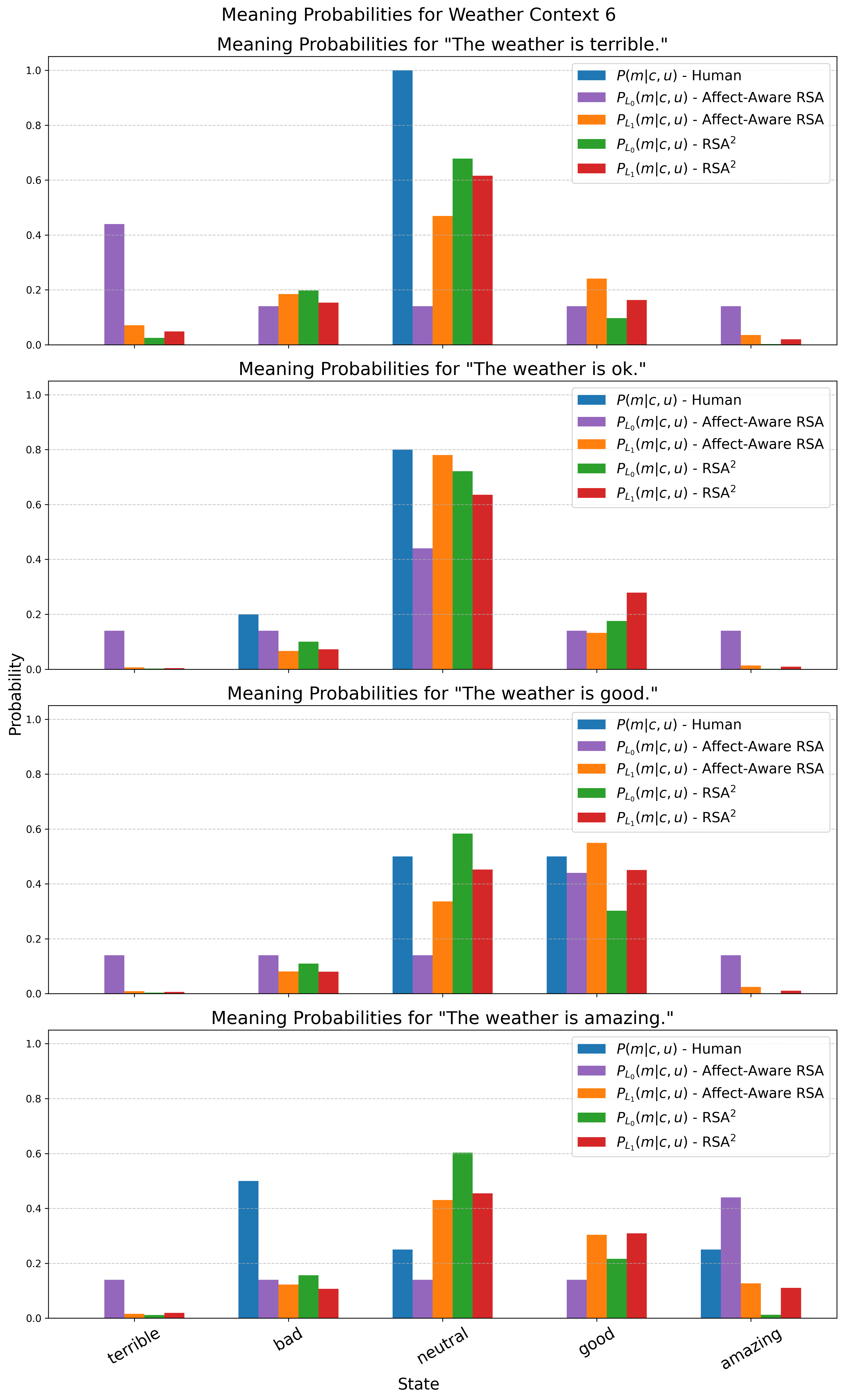}
    \caption{Meaning probability distributions by humans along with the listener and pragmatic listeners of both affect-aware RSA and \rsatwospace for weather context 6 from Fig.~\ref{fig:weather_images}.}
    \label{fig:Weather_context_6}
\end{figure}

\begin{figure}
    \centering
    \includegraphics[width=0.6\linewidth]{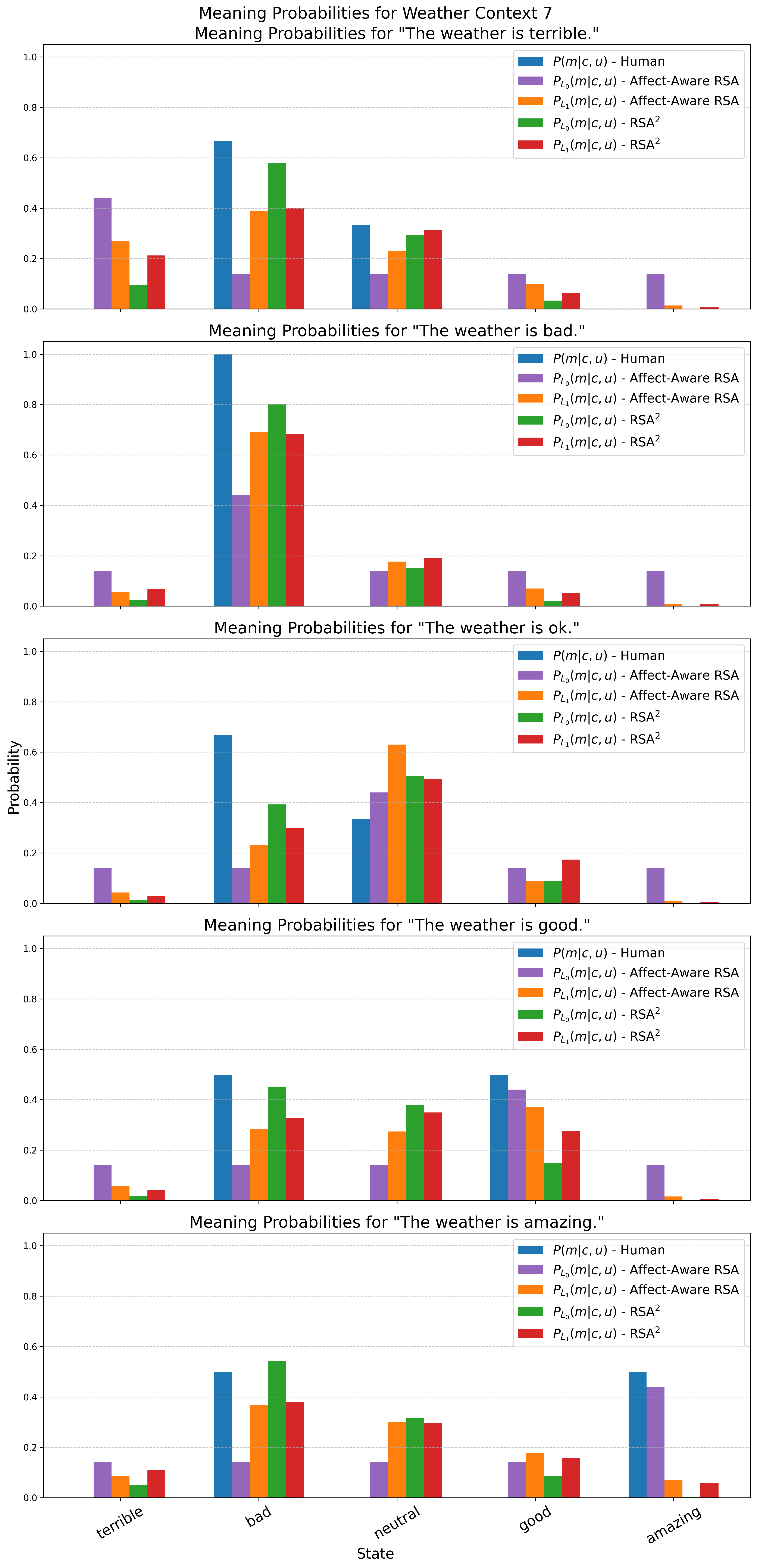}
    \caption{Meaning probability distributions by humans along with the listener and pragmatic listeners of both affect-aware RSA and \rsatwospace for weather context 7 from Fig.~\ref{fig:weather_images}.}
    \label{fig:Weather_context_7}
\end{figure}

\begin{figure}
    \centering
    \includegraphics[width=0.6\linewidth]{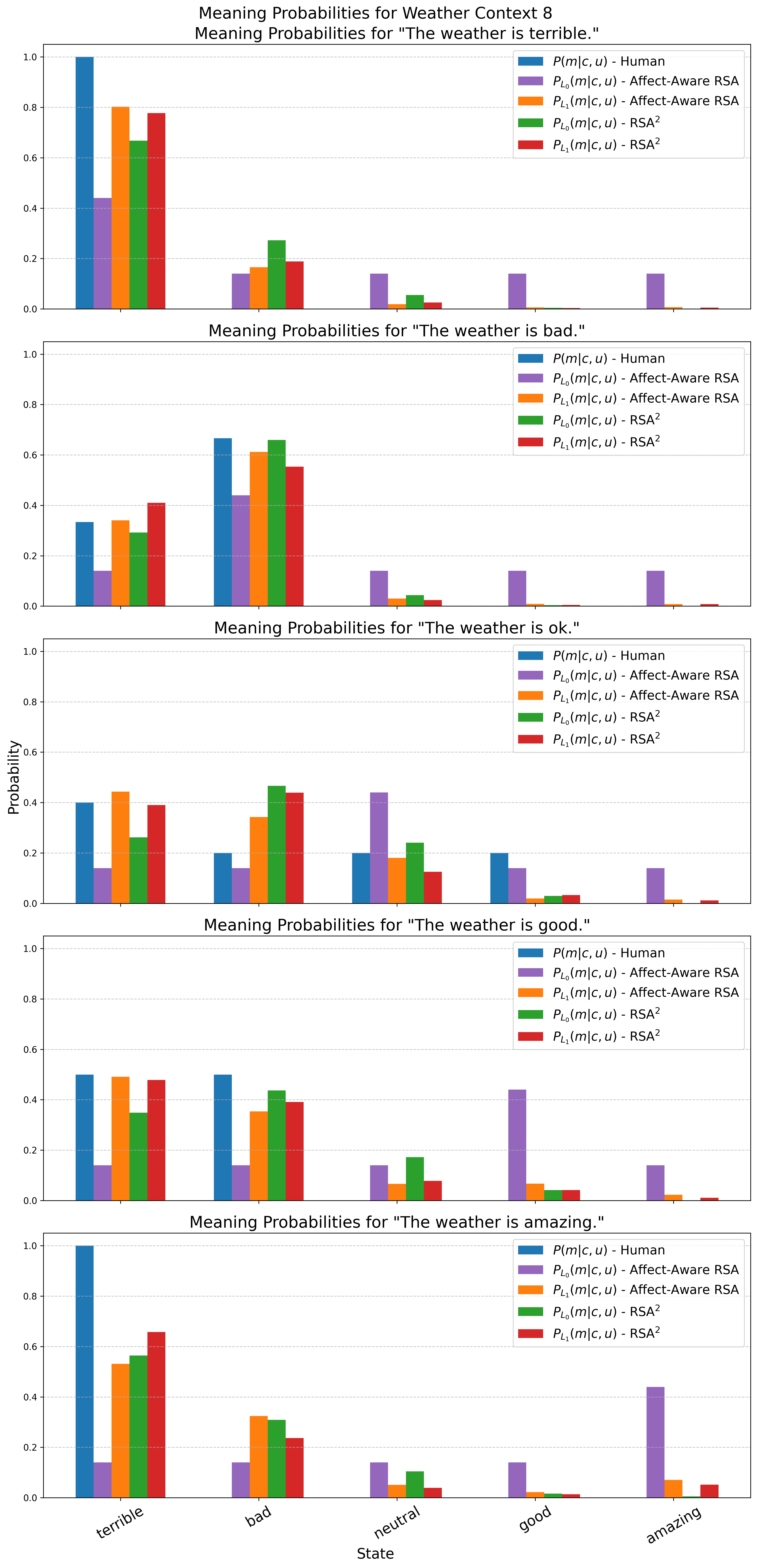}
    \caption{Meaning probability distributions by humans along with the listener and pragmatic listeners of both affect-aware RSA and \rsatwospace for weather context 8 from Fig.~\ref{fig:weather_images}.}
    \label{fig:Weather_context_8}
\end{figure}

\begin{figure}
    \centering
    \includegraphics[width=0.6\linewidth]{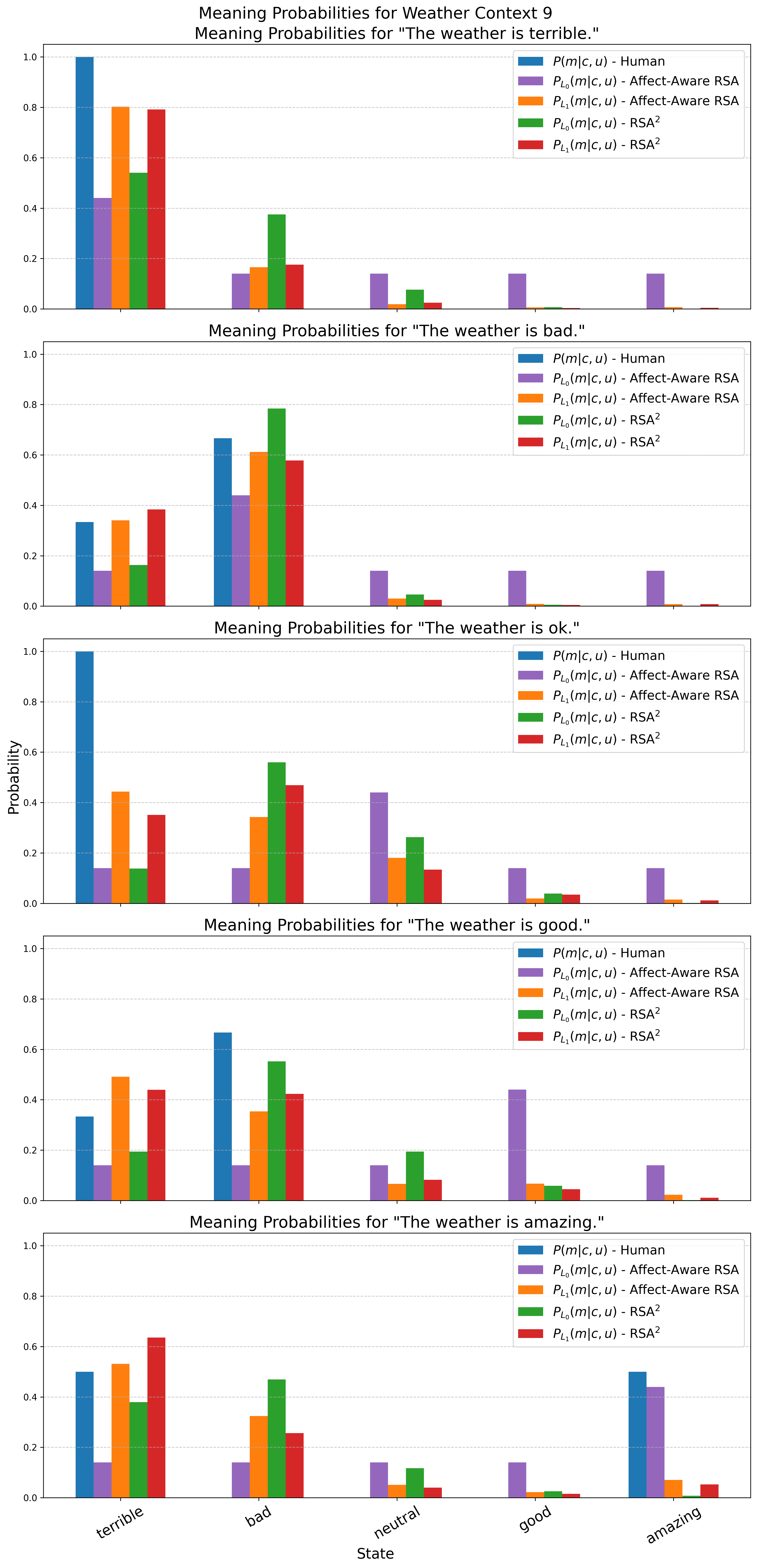}
    \caption{Meaning probability distributions by humans along with the listener and pragmatic listeners of both affect-aware RSA and \rsatwospace for weather context 9 from Fig.~\ref{fig:weather_images}.}
    \label{fig:Weather_context_9}
\end{figure}

Meaning probability distributions by humans, alongside the listener and pragmatic listeners of both affect-aware RSA and \rsatwospace models, are presented for each of the nine weather contexts (which are associated with Fig.~\ref{fig:weather_images}). Figures \ref{fig:Weather_context_1}, \ref{fig:Weather_context_2}, and \ref{fig:Weather_context_3} illustrate these distributions for weather contexts 1, 2, and 3 where the weather is visibly good. Figures \ref{fig:Weather_context_4}, \ref{fig:Weather_context_5}, and \ref{fig:Weather_context_6} show the distributions for weather contexts 4, 5, and 6 where the weather is neither visibly good nor bad. Finally, distributions for weather contexts 7, 8, and 9 where the weather is visibly bad are shown in Figures \ref{fig:Weather_context_7}, \ref{fig:Weather_context_8}, and \ref{fig:Weather_context_9}.

\newpage

\section{LLM Irony Interpretation with \rsatwo}

\label{app:pragmega_plus_exp}

\subsection{PragMega+ Dataset Examples}

\label{app:pragmega_more_examples}

\begin{table*}[ht]
    \centering
    \resizebox{\linewidth}{!}{
    \begin{tabular}{p{6cm}p{6cm}p{6cm}}
    \toprule
    Non-literal Intended Meaning & Literal Intended Meaning & Intended Meanings \\
    \midrule
    In a shop, Lara tries on a dress. \textbf{The dress is far too long for her.} Lara asks Simon: ``Does this dress fit me?'' Simon answers: ``Wow! That must be custom made! It's clearly the perfect size and length for you.'' & In a shop, Lara tries on a dress. \textbf{The dress fits her perfectly.} Lara asks Simon: ``Does this dress fit me?'' Simon answers: ``Wow! That must be custom made! It's clearly the perfect size and length for you.'' & 1. The dress is fitting well. (\texttt{LM}) \newline 2. Lara needs to get a dress of a shorter length. (\texttt{NLM}) \newline 3. Simon does not like the color of this dress. (\texttt{OM}) \newline 4. Simon has to get back to work. (\texttt{NSM}) \\
    \midrule
    While Tom and a new acquaintance from work, Sara, were chatting at a party, they noticed a colleague across the room. \textbf{She was standing alone holding a drink and a CD.} Tom points at the girl and comments:  ``The life of the party, right there.'' & While Tom and a new acquaintance from work, Sara, were chatting at a party, they noticed a colleague across the room. \textbf{She was standing at the center of a group of colleagues, holding a drink and and telling another one of her classic stories.} Tom points at the girl and comments: ``The life of the party, right there.''  & 1. Their colleague is very sociable. (\texttt{LM}) \newline 2. Their colleague is quite unsociable. (\texttt{NLM}) \newline 3. Their colleague has good taste in music. (\texttt{OM}) \newline 4. Their colleague has good taste in fashion. (\texttt{NSM}) \\
    \midrule
    After a long day, Bruce returns home and notices \textbf{that his kids have not cleaned their room.} Bruce says ``I love how clean your room is.'' What did Bruce want to convey? & After a long day, Bruce returns home and notices \textbf{that his kids have cleaned their room.} Bruce says ``I love how clean your room is.'' What did Bruce want to convey? & 1. Bruce is happy with how clean his kids' room is. (\texttt{LM}) \newline 2. Bruce is annoyed that his kids have not cleaned their room. (\texttt{NLM}) \newline 3. It's important to keep one's room clean. (\texttt{OM}) \newline 4. Bruce forgot to make dinner. (\texttt{NSM}) \\
    \midrule
    After asking his parents several times, Edward has finally received a new gaming console for his birthday. \textbf{However, even though Edward promised that he would not spend more than two hours a day gaming, his parents quickly realize that Edward has no intention of keeping that promise.} Edward's mom tells he, ``I see you are quite good at keeping your promises.'' & After asking his parents several times, Edward has finally received a new gaming console for his birthday. \textbf{Edward promised that he would not spend more than two hours a day gaming.} To their surprise, Edward's parents realize that Edward has every intention of keeping that promise. Edward's mom tells he, ``I see you are quite good at keeping your promises.'' & 1. She is disappointed that Edward has not kept he promise. (\texttt{LM}) \newline 2. Edward has kept his promise. (\texttt{NLM}) \newline 3. Spending too much time gaming is bad for one's health. (\texttt{OM}) \newline 4. She needs to buy a new remote control. (\texttt{NSM}) \\
    \bottomrule
    \end{tabular}
    }
    \caption{Examples of a scenario with a non-literal intended meaning as well as its corresponding modified scenario with a literal intended meaning in the PragMega+ dataset as well as the shared intended meanings: 1. \texttt{Literal Meaning} (\texttt{LM})  2. \texttt{Non-Literal Meaning} (\texttt{NLM}) 3. \texttt{Overlap Meaning} (\texttt{OM}) 4. \texttt{Non-Sequitur Meaning} (\texttt{NSM}). The first two scenarios are from the original PragMega dataset while the last two were manually generated by the authors as part of the test set.}
    \label{tab:pragmega_examples}
\end{table*}

\subsection{Prompt Templates}

\label{app:prompts}

We designed four prompt templates to elicit different conditional meaning distributions from language models. Each prompt targets a specific probability of interest within our framework. Figure~\ref{fig:prompt_p_mcu} elicits $P(m \mid c, u)$, Figure~\ref{fig:prompt_p_rcu} elicits $P(r \mid c, u)$, Figure~\ref{fig:prompt_p_mcur} elicits for $P(m \mid c, u, r)$, and Figure~\ref{fig:prompt_p_mc} elicits $P(m \mid c)$.

\begin{figure}[htbp]
    \centering
    \begin{tcolorbox}[
        title={Prompt Template for $P(m|c,u)$},
        colback=gray!10!white, 
        colframe=gray!50!black, 
        coltitle=white, 
        fonttitle=\bfseries, 
        boxsep=5pt, 
        arc=4pt, 
        boxrule=0.8pt, 
        left=10pt, 
        right=10pt, 
        top=5pt, 
        bottom=5pt 
    ]
    Template:
    \begin{verbatim}
    Task: You will read short stories that describe everyday situations. 
    Each story will be followed by a multiple-choice question.
    Read each story and choose the best answer.
    Your task is to decide what the character in the story is trying to convey.
    The answer options are 1, 2, 3, or 4.
    
    
    [scenario] 
    What meaning is X likely conveying?
    
    [options]
    
    Answer:
    \end{verbatim}
    Sample:
    \begin{verbatim}
    Task: You will read short stories that describe everyday situations.
    Each story will be followed by a multiple-choice question. 
    Read each story and choose the best answer. 
    Your task is to decide what the character in the story is trying to convey.
    The answer options are 1, 2, 3, or 4.
    
    
    John is a teacher at an elementary school. When talking with the
    principal about a new student, who did poorly on her entrance
    examination, John said, ``This one is really sharp.'' 
    What meaning is John trying to convey?
    
    1. The pencils need to be sharpened.
    2. The student is smart.
    3. The student is not very clever.
    4. The entrance exam is unfair. 
    
    Answer:
    \end{verbatim}
    \end{tcolorbox}
    \caption{Prompt template for $P(m|c,u)$ where [scenario] is replaced with one of the (ironic or literal) scenarios from the PragMega dataset with the utterance and [options] are the 4 possible intended meanings for this particular scenario. $X$ is replaced with the name of the speaker in the scenario which is extracted using a regular expression.}
    \label{fig:prompt_p_mcu}
\end{figure}

\begin{figure}[htbp]
    \centering
    \begin{tcolorbox}[
        title={Prompt Template for $P(r|c,u)$},
        colback=gray!10!white, 
        colframe=gray!50!black, 
        coltitle=white, 
        fonttitle=\bfseries, 
        boxsep=5pt, 
        arc=4pt, 
        boxrule=0.8pt, 
        left=10pt, 
        right=10pt, 
        top=5pt, 
        bottom=5pt 
    ]
    Template:
    \begin{verbatim}
    Task: You will read short stories that describe everyday situations.
    Each story will be followed by a multiple-choice question. 
    Read each story and choose the best answer. 
    Your task is to decide, given the situation and what the character has said,
    whether the character in the story is being "Sincere" or "Not Sincere".
    A character is "Sincere" if what they are saying is consistent with the
    context of the story. For instance, if the weather outside is 
    sunny then a "Sincere" utterance might be "The weather 
    outside is amazing." Likewise, if the weather outside is rainy then
    a "Sincere" utterance might be "The weather is terrible." 
    In contrast, a character is "Not Sincere" if they are saying something
    that contradicts the context of the story. For instance, if the weather 
    outside is rainy then a "Not Sincere" utterance might be "The weather 
    outside is amazing." Similarly, if the weather outside is sunny then 
    a "Not Sincere" utterance might be "The weather outside is terrible." 
    The answer options are 1 or 2.
    
    [scenario] 
    Was X being sincere or not sincere?
    
    [options]
    
    Answer:
    \end{verbatim}
    Sample:
    \begin{verbatim}
    Task: You will read short stories that describe everyday situations.
    Each story will be followed by a multiple-choice question. 
    Read each story and choose the best answer. 
    Your task is to ... [see above].
    The answer options are 1 or 2.
    
    
    John is a teacher at an elementary school. When talking with the
    principal about a new student, who did poorly on her entrance
    examination, John said, ``This one is really sharp.'' 
    Was John being sincere or not sincere?
    
    1. Sincere 
    2. Not Sincere
    
    Answer:
    \end{verbatim}
    \end{tcolorbox}
    \caption{Prompt template for $P(r|c,u)$ where [scenario] is replaced with one of the (ironic or literal) scenarios from the PragMega dataset with the utterance and [options] are either sincere or not sincere. $X$ is replaced with the name of the speaker in the scenario which is extracted using a regular expression. We used the ``Sincere''/``Not sincere'' terminology because we found that the term ``Irony'' would hurt performance.}
    \label{fig:prompt_p_rcu}
\end{figure}


\begin{figure}[htbp]
    \centering
    \begin{tcolorbox}[
        title={Prompt Template for $P(m|c,u,r)$},
        colback=gray!10!white,
        colframe=gray!50!black,
        coltitle=white,
        fonttitle=\bfseries,
        boxsep=5pt,
        arc=4pt,
        boxrule=0.8pt,
        left=10pt,
        right=10pt,
        top=5pt,
        bottom=5pt
    ]
    Template:
    \begin{verbatim}
Task: You will read short stories that describe everyday situations and which 
finish with a character saying something. Your task is to decide, given the 
situation and what the character has said, what meaning the character is trying 
to convey. Each story will be followed by 4 possible meaning interpretations 
listed from 1 to 4. Read each story and choose the number corresponding to the 
best meaning interpretation. You can only answer with 1, 2, 3, or 4.
    \end{verbatim}

    If sincere:
    \begin{verbatim}
Assume that the character is saying exactly what they want to convey literally 
when choosing the character's true intended meaning **even if** it contradicts 
the context. For instance, if it's a sunny day and the character says "The 
weather is terrible." then they do actually mean that the weather is terrible. 
Similarly, if it's a rainy day and the character says "The weather is amazing." 
then they do actually mean that the weather is amazing.
    \end{verbatim}

    If not sincere: 
    \begin{verbatim}
In addition, when choosing the intended meaning, assume that the character is 
saying the opposite of what they want to convey. For instance, if they say "The 
weather is terrible." then they actually mean that the weather is amazing. 
Similarly, if they say "The weather is amazing." then they actually mean that 
the weather is terrible. 
    \end{verbatim}

    \begin{verbatim}
[scenario] 
What meaning is X trying to convey?

[options]

Answer:
    \end{verbatim}
    \end{tcolorbox}
    \caption{Prompt template for $P(m|c,u,r)$ where [scenario] is replaced 
    with one of the (ironic or literal) scenarios from the PragMega dataset 
    with the utterance and [options] are the 4 possible intended 
    meanings for this particular scenario. $X$ is replaced with the name of 
    the speaker in the scenario which is extracted using a regular expression.}
    \label{fig:prompt_p_mcur}
\end{figure}

\begin{figure}[htbp]
    \centering
    \begin{tcolorbox}[
        title={Prompt Template for $P(m|c)$},
        colback=gray!10!white, 
        colframe=gray!50!black, 
        coltitle=white, 
        fonttitle=\bfseries, 
        boxsep=5pt, 
        arc=4pt, 
        boxrule=0.8pt, 
        left=10pt, 
        right=10pt, 
        top=5pt, 
        bottom=5pt 
    ]
    Template:
    \begin{verbatim}
    Task: You will read short stories that describe everyday situations 
    and which finish with a character saying something. Your task is to 
    decide, given the situation, what meaning the character is most 
    likely conveying. Each story will be followed by 4 possible meaning 
    interpretations listed from 1 to 4. Read each story and choose the 
    number corresponding to the most likely meaning. You can only answer 
    with 1, 2, 3, or 4.

    [scenario] 
    What meaning is X likely conveying?
    
    [options]
    
    Answer:
    \end{verbatim}
    Sample:
    \begin{verbatim}
    Task: You will read short stories that describe everyday situations 
    and which finish with a character saying something. Your task is ... [see above]
    
    
    John is a teacher at an elementary school. When talking with the
    principal about a new student, who did poorly on her entrance
    examination, John said something.
    What meaning is John likely conveying?
    
    1. The pencils need to be sharpened.
    2. The student is smart.
    3. The student is not very clever.
    4. The entrance exam is unfair. 
    
    Answer:
    \end{verbatim}
    \end{tcolorbox}
    \caption{Prompt template for $P(m|c)$ where [scenario] is replaced with one of the (ironic or literal) scenarios from the PragMega dataset without the utterance and [options] are the 4 possible intended meanings for this particular scenario. $X$ is replaced with the name of the speaker in the scenario which is extracted using a regular expression.}
    \label{fig:prompt_p_mc}
\end{figure}

\newpage

\subsection{Additional Results}

\begin{table}[h]
\centering
\resizebox{\linewidth}{!}{
\begin{tabular}{p{0.22\textwidth} 
                 >{\centering\arraybackslash}p{0.18\textwidth} 
                 >{\centering\arraybackslash}p{0.18\textwidth}
                 >{\centering\arraybackslash}p{0.18\textwidth}
                 >{\centering\arraybackslash}p{0.18\textwidth}}
\toprule
& \makecell[c]{$P_{L_0}(m = \text{\texttt{NLM}} \mid$ \\ $c,u,r=\text{\textit{irony}})$}
& \makecell[c]{$P_{L_1}(m = \text{\texttt{NLM}} \mid$ \\ $c,u,r=\text{\textit{irony}})$}
& \makecell[c]{$P_{L_0}(m = \text{\texttt{LM}} \mid$ \\ $c,u,r=\text{\textit{literal}})$}
& \makecell[c]{$P_{L_1}(m = \text{\texttt{LM}} \mid$ \\ $c,u,r=\text{\textit{literal}})$} \\
\midrule
\makecell[c]{Scenarios where the\\intended meaning is\\non-literal} & 0.92 & 0.94 & 0.47 & 0.62 \\
\makecell[c]{Scenarios where the\\intended meaning is\\literal}      & 0.16 & 0.59 & 0.95 & 0.88 \\
\bottomrule
\end{tabular}

}
\caption{Average listener probability distributions \emph{conditioned on} the \textit{irony} and \textit{literal} rhetorical strategies on both the ironic split (i.e., the split in which the intended meaning is non-literal) and the literal split (i.e., the split in which the intended meaning is literal).}
\label{tab:listeners_by_rs}
\end{table}

\begin{table}[h]
\centering
\begin{tabular}{lcc}
\toprule
& $P_N(r = \textit{irony} \mid c, u)$ 
& $P_N(r = \textit{literal} \mid c, u)$ \\
\midrule
\makecell[c]{Scenarios where the\\intended meaning is\\non-literal} & 0.88 & 0.12 \\
\makecell[c]{Scenarios where the\\intended meaning is\\literal}     & 0.45 & 0.55 \\
\bottomrule
\end{tabular}
\caption{Rhetorical strategy posteriors $P_N(r|c,u)$ on both the ironic split (i.e., the split in which the intended meaning is non-literal) and the literal split (i.e., the split in which the intended meaning is literal).}
\label{tab:p_r_c_u}
\end{table}

\newpage

\twocolumn

\section{\rsatwolearnedspace: A Rhetorical Strategy Clustering Algorithm}

\label{app:rsc_rsa}

\begin{algorithm}
\caption{\rsatwolearned}\label{alg:learned_rsa}
\begin{algorithmic}[1]
\Ensure Context $c$, Meaning $m$, Utterance $u$
\Require Set of induced rhetorical function values $F(c, m, u) = \{f_1(c, m, u), \dots, f_n(c, m, u)\}$

\Comment{Initialize the following objects.}

\State LLMs $M$ and $G$
\State Embedding function $\texttt{embed}: \mathcal{U} \to \mathbb{R}^n$
\State K-means clustering function $\texttt{k-means}: \mathcal{P}(\mathbb{R}^n) \to \mathcal{P}(\mathbb{R}^n)$
\State Cosine similarity $\texttt{cosine-sim}: \mathbb{R}^n \times \mathbb{R}^n \to [0,1]$

\Comment{Generate, embed and cluster alternative utterances.}
 
\State $\mathcal{U}_\text{alt} \leftarrow \{ u \} \cup \{ u_i \sim P_G(u|c) \}$ 
\State $E \leftarrow \{ \texttt{embed}(u') : u' \in \mathcal{U}_\text{alt} \}$
\State $\mathcal{X} \leftarrow \texttt{k-means}(E)$

\Comment{Compute rhetorical function value for each cluster.}

\State $F(c, m, u) \leftarrow \emptyset$
\For{$r \in \mathcal{X}$ }
    \State $\mathcal{U}_r \leftarrow \{u' \in \mathcal{X}_\text{alt} : u' \in \text{ cluster }r\}$
    
    \Comment{Compute cluster meaning probability and weight it by the utterance distance to the centroid.}
    
    \State $p_{mc}\! \leftarrow \! \frac{\sum_{u' \in \mathcal{U}_r}P_{M}(m|c,u')}{\sum_{m' \in \mathcal{M}}\sum_{u' \in \mathcal{U}_r}P_{M}(m'|c,u')}$
    \State $f_r(c, m, u)\! \leftarrow \! \frac{p_{mc}}{P_M(m|c,u)}$
    \State Append $f_r(c, m,u)$ to $F(c, m, u)$
\EndFor

\State \Return $F(c, m, u)$ 
\end{algorithmic}
\end{algorithm}
We present our clustering algorithm, \textbf{R}hetorical \textbf{S}trategy \textbf{C}luster RSA \rsatwolearned, which attempts to automatically \emph{induce} the most salient rhetorical strategies and their corresponding rhetorical functions. To do so, we rely on the intuition that utterances which are generated with the same intended meaning and rhetorical strategy are likely to be semantically similar. For example, the utterances ``The weather is amazing.'' and ``Gosh, this weather is so great!'', uttered in the context of bad weather, are likely to both employ the same rhetorical strategy (irony) to convey the same intended meaning (that the weather is in fact terrible). The algorithm which computes the set of induced rhetorical function values, $F(c, m, u) = \{f_1(c, m, u), \dots, f_n(c, m, u)\}$, generates alternative utterances using a base LLM, embeds them and clusters them with k-means to induce \emph{rhetorical strategy clusters} which act as proxies of prototypical rhetorical strategies. The full procedure is presented in Algorithm~\ref{alg:learned_rsa}.

In brief, the \rsatwolearnedspace algorithm generates, embeds and clusters a set of alternative utterances which could have occurred in context $c$ to create the rhetorical strategy clusters $\mathcal{X}$. Using the generated clusters $\mathcal{X}$, \rsatwolearnedspace computes the corresponding rhetorical function values, $f_r(c, m, u)$, for each cluster $r$ for a given $c,m,u$ triple. To do so, \rsatwolearnedspace uses a formula which averages the meaning probabilities of all the utterances in that cluster and divides them by $P_M(m|c,u)$. In this way, the induced rhetorical strategy values parallel those from Equation~\ref{eq:what_is_frcmu} in that they return a ratio of two probabilities. The equations for computing $f_r$ are as follows:%
\begin{align}
    p_{mc} &= \frac{\sum_{u' \in \mathcal{U}_r}P_{M}(m|c,u')}{\sum_{m' \in \mathcal{M}}\sum_{u' \in \mathcal{U}_r}P_{M}(m'|c,u')}, \\
    f_r(c,m,u) &= \frac{p_{mc}}{P_M(m|c,u)}.
\end{align}%

These values are then used to compute $P_{L_0}(m|c,u,r)$ as defined in Equation~\ref{eqn:rsatwol0}. To marginalize across rhetorical strategy clusters, we use the relative cluster size, i.e. $P_{R|CU}(r|c,u) = \frac{|\mathcal{U}_r|}{|\mathcal{U}|}$.

If the alternative utterances for $u = \text{``The weather is amazing.''}$ include $u_1 = \text{``Gosh, this weather is so great!''}$, $u_2 = \text{``The weather is terrible.''}$ and $u_3 = \text{``The weather is so bad.''}$, then we expect for the embeddings of $u$ and $u_1$ to be in one cluster and for the embeddings of $u_2$ and $u_3$ to be in another cluster. These two rhetorical strategy clusters would then resemble the \emph{ironic} and \emph{literal} rhetorical strategies from Section~\ref{sec:rsatwooriginal} and their \rsatwolearned-derived rhetorical function values would approximate those of $f_\textit{irony}$ for $u$ and $u_1$ and of $f_\textit{literal}$ for $u_2$ and $u_3$.

\begin{table*}[ht]
    \centering
    \resizebox{\linewidth}{!}{
    \begin{tabular}{>{\raggedright\arraybackslash}p{0.25\textwidth}
                    >{\centering\arraybackslash}p{0.22\textwidth}
                    >{\centering\arraybackslash}p{0.22\textwidth}
                    >{\centering\arraybackslash}p{0.22\textwidth}}
        \toprule
        Listener & 
        \makecell[c]{Average $P(m|c,u)$ \\ across all scenarios} & 
        \makecell[c]{Average $P(m|c,u)$ \\ across ironic scenarios} & 
        \makecell[c]{Average $P(m|c,u)$ \\ across literal scenarios} \\
        \midrule
        RSA-RSC – $L_0$ with $P(m|c)$ & 0.59 & 0.91 & 0.28 \\
        RSA-RSC – $L_1$ with $P(m|c)$ & 0.66 & 0.91 & 0.41 \\
        RSA-RSC – $L_0$ without $P(m|c)$ & 0.27 & 0.50 & 0.034 \\
        RSA-RSC – $L_1$ without $P(m|c)$ & 0.28 & 0.52 & 0.036 \\
        \bottomrule
    \end{tabular}
    }
    \caption{Average $P(m|c,u)$ for different RSA-RSC listener models.}
    \label{tab:rsa_rsc_clustering_performance}
\end{table*}

\subsection{\rsatwolearnedspace Implementation Details}

To implement the embedding and clustering procedures of \rsatwolearned, we utilized the Sentence-BERT architecture \citep{reimers-gurevych-2019-sentence} and the k-means clustering algorithm from scikit-learn \citep{sklearn_api}. The k-means algorithm was executed with 10 initializations, using default settings from scikit-learn.

\subsection{Clustering Results}

We experimented with 2, 4, and 8 clusters yielding no significant difference. As a result, we report results of the \rsatwolearnedspace clustering method with 4 clusters with and without the meaning prior, $P(m|c)$, in Table~\ref{tab:rsa_rsc_clustering_performance}. We see that across all listeners and scenario types, the performance is largely driven by the meaning prior $P(m|c)$ with the performance on the literal scenarios being worse than random. We encourage future work to improve upon our initial attempt at automatically uncovering the most salient rhetorical strategies in a given context and leveraging them within the \rsatwospace framework.

\end{document}